%% file: main_camera_ready.tex
\documentclass{article}

\newif\ifincludechecklist
\includechecklistfalse

\PassOptionsToPackage{numbers,compress}{natbib}
\usepackage[eandd, final]{neurips_2026}

\usepackage[utf8]{inputenc}
\usepackage[T1]{fontenc}
\usepackage{microtype}
\usepackage{xcolor}
\usepackage{hyperref}
\usepackage{url}
\usepackage{fontawesome5}
\usepackage{amsmath,amssymb}
\usepackage{graphicx}
\usepackage{booktabs}
\usepackage{array,tabularx}
\usepackage{wrapfig}
\usepackage{tcolorbox}
\usepackage[font=small,labelfont=bf]{caption}

\hypersetup{
    colorlinks=true,
    citecolor=blue,
    linkcolor=black,
    urlcolor=teal
}

\title{BabyTheorist: A Benchmark for Learning to Theorize the World from Observation Alone}

\author{%
  \hspace*{-\tabcolsep}\makebox[\textwidth][c]{%
    \textbf{Doojin Baek}$^{*\dagger}$ \quad
    \textbf{Junyeob Baek}$^{*\dagger}$ \quad
    \textbf{Mingyu Jo}%
  } \\
  \vspace{2mm} \\
  \hspace*{-\tabcolsep}\makebox[\textwidth][c]{%
    \textbf{Hosung Lee} \quad
    \textbf{Taegu Kang} \quad
    \textbf{Gyubin Lee} \quad
    \textbf{Sungjin Ahn}%
  } \\
  \vspace{4mm} \\
  \hspace*{-\tabcolsep}\makebox[\textwidth][c]{KAIST} \\
}

\begin{document}
\maketitle
{\renewcommand{\thefootnote}{}%
\footnotetext{$^{*}$Equal contribution}%
\footnotetext{$^{\dagger}$Correspondence to: Doojin Baek (\texttt{doojin.a.baek@kaist.ac.kr}) and Junyeob Baek (\texttt{wnsdlqjtm@kaist.ac.kr})}}

\vspace{-2em}
\begin{center}
  \faGithub\ \textbf{Code}: \href{https://github.com/ahn-ml/babytheorist}{\texttt{github.com/ahn-ml/babytheorist}}\\
  \faDatabase\ \textbf{Dataset}: \href{https://huggingface.co/datasets/ahn-ml/babytheorist}{\texttt{huggingface.co/datasets/ahn-ml/babytheorist}}
\end{center}
\vspace{0.5em}

\begin{abstract}
How would we know whether a machine truly understands the world? We argue that one key capability is \textit{observational theory learning}: acquiring reusable theories from pure observations and applying them to new situations. Yet current visual reasoning benchmarks, such as ARC-AGI, mainly evaluate the latter, presenting each task as a few-shot puzzle grouped by a common rule. Whether models acquire reusable theories from observations remains untested. We introduce BabyTheorist, a benchmark that brings this acquisition under controlled training and evaluation. BabyTheorist generates continual streams of before-after visual observations from hidden compositional programs; during training, learners see only the resulting transitions, with no program labels, task identities, or support sets. To support theory learning and evaluation, BabyTheorist combines three design controls---a hidden program language, frequent subprograms, and a curriculum over program complexity---with a continual train-test protocol featuring level skipping. Experiments with seven representative baselines show that BabyTheorist poses challenges that current program-learning approaches have yet to resolve.
\end{abstract}

\section{Introduction}

Long before they learn to solve problems, children learn how the world works. They watch ordinary phenomena unfold---objects falling, water freezing, people acting---and gradually accumulate reusable theories that explain them. Faced with a novel situation, they do not start from scratch but recombine theories they have already built. This suggests two distinct capabilities behind everyday understanding: \textbf{theory acquisition} from raw observation, and \textbf{theory application}, which recombines acquired theories to explain new situations (Figure~\ref{fig:observational_learning}).

Current visual reasoning benchmarks focus on the second. ARC-AGI~\citep{chollet2019measure, arc24, arc25}, Bongard problems~\citep{nie2021bongardlogo, jiang2023bongardhoi}, and Raven-style suites~\citep{zhang2019raven} each present a task as a pre-grouped support set sharing a hidden rule, then test whether a model can infer the rule and apply it to a query. This setting is valuable, but the process preceding such rule inference---accumulating reusable and composable primitives and rules from many \emph{ungrouped} observations---is left outside the evaluation. Thus, a model can be scored on few-shot transfer without being asked how it acquired the theories behind that transfer.

\input{CameraReady/figures/fig_sec1_observational_learning_general}
We argue that progress toward \textbf{observational theory learning}---acquiring reusable theories from observations and applying them to new situations---requires closer study of \emph{theory acquisition} from ungrouped observations as experience accumulates. Building a benchmark for this phase is the central goal of this paper. We further commit to a \emph{learning-focused} stance: rather than giving learners a predefined domain-specific language (DSL) or primitive vocabulary for symbolic search, our setup asks whether neural learners can discover the relevant primitives and theories directly from raw observations.

\textbf{From paradigm to benchmark.}
Studying observational theory learning first requires a formal definition. The \textit{Learning-to-Theorize} (L2T) paradigm~\citep{baek2026l2t} considers a learner that observes a scene before a change, $x$, and after it, $y$, without knowing which operations transformed one into the other. Each such observation pair $(x,y)$ is called a \emph{phenomenon}. To explain the observed change, the learner must discover reusable operations, called \emph{primitives}, and infer how they were composed to transform $x$ into $y$. An ordered sequence of these primitives forms a \emph{program} $\tau$. Neither the primitives nor the programs are provided; the learner receives only raw $(x, y)$ pairs, without program annotations, action labels, or task groupings. This commitment is intentionally minimal—any pair of observations connected by a hidden mechanism is a valid training example, opening the door to uncurated observational data at scale. However, L2T's models were trained on static batch datasets with programs of depth 1--3, leaving \emph{theory acquisition over experience}---building and updating a theory system as phenomena grow---out of scope.

\textbf{BabyTheorist.}
We close this gap with \textbf{BabyTheorist}, a benchmark that extends L2T's observational learning setting from static batch datasets to a continual curriculum of increasingly complex phenomena. It provides three design controls: (i) a \emph{hidden DSL} defines the primitives and their composition rules without revealing them to the learner; (ii) \emph{frequent subprograms} introduce recurring sequences of primitives that learners can discover and reuse; and (iii) a \emph{level-based curriculum} introduces increasingly complex compositions. 
Building on these controls, our continual train--test protocol is designed to reward the acquisition of reusable theories. Each level begins with an entry evaluation; learners that pass skip that level's training observations (\emph{level skipping}), testing whether their acquired theories transfer to increasingly complex phenomena.

\textbf{Where existing methods stand.}
To characterize what BabyTheorist measures, we evaluate seven representative baselines: black-box predictors (LLMs, autoregressive Transformers), monolithic latent encoders (Cont-Mono, Disc-Mono) and their variants (Cont-Mono-Opt, Cont-Mono-OC), and the compositional discrete-program model NEO~\citep{baek2026l2t}. All baselines stall within the first two of the eight curriculum levels. Analyzing their errors, we identify three challenges (Sections~\ref{sec:challenge_primitives}--\ref{sec:exp_compositional}): learners struggle to \emph{discover reusable primitives}, to \emph{learn where to apply} a transformation in multi-object scenes, and to \emph{explain higher-depth phenomena} that combine several primitives.

\textbf{Contributions.} The main contributions of this paper are as follows:
\begingroup\setlength{\leftmargini}{1.2em}\setlength{\topsep}{0pt}\setlength{\partopsep}{0pt}
\begin{itemize}\setlength{\itemsep}{0pt}\setlength{\parskip}{0pt}
    \item \textbf{A learning-focused benchmark.} BabyTheorist evaluates theory acquisition from ungrouped observations and application to new tasks throughout a continual curriculum.
    \item \textbf{Three design controls.} A hidden DSL, frequent subprograms, and a level-based curriculum enable controlled study of observational theory learning across four benchmark modes.
    \item \textbf{A continual evaluation protocol.} We design a continual train--test protocol with level skipping to reward theory reuse as program complexity increases, ranking learners by \textsc{Score} and using \textsc{Efficiency} to break ties in favor of earlier progress.
    \item \textbf{An empirical assessment.} Seven representative baselines show limited progress through the curriculum, revealing challenges that existing approaches have yet to resolve.
\end{itemize}
\endgroup

\section{Learning to Theorize the World Just by Seeing}
\label{sec:preliminaries}
\input{CameraReady/figures/fig_sec1_overview}
To formalize observational theory learning (Figure~\ref{fig:overview}), we build on \textbf{Learning-to-Theorize}~\citep{baek2026l2t}, which models the hidden mechanism behind each observation pair as a latent compositional program.

\label{sec:l2t}

\textbf{Theorize Just by Seeing.}
Imagine watching a stream of (before, after) image pairs in which a hidden rule transforms each before-image into its after-image. A standard predictor scored on reproducing the after-image can succeed without ever isolating the rule. We ask a different question: \emph{Can a learner, just by watching, recover the hidden mechanisms behind what it sees and apply them to situations it has never encountered?} L2T~\citep{baek2026l2t} answers by treating each (before, after) pair as a \emph{phenomenon} and the rule behind it as a \emph{program}: a short sequence of primitive operations whose execution reproduces the observed transformation. The learned primitives form a vocabulary of reusable operations, akin to the ``mental programs'' of compositional cognition~\citep{chater2013programs,dehaene2022symbols}.

\textbf{Phenomena and programs.}
\label{phenomena_and_programs}
Let $\mathcal{X}$ denote the space of observations. A \textbf{phenomenon} is an observation pair $(x,y)$, where $x \sim p(x)$ is a source observation and $y$ is a target observation generated by an unobserved program $\tau$ that transforms $x$ into $y$. Following L2T, we represent each program as a sequence of primitive operations, each transforming one observation into another. Let $\mathcal{Z} = \{z_1,\ldots,z_M\}$ denote the set of $M$ primitives, with each primitive $z_i$ associated with an execution function $f_{z_i} : \mathcal{X} \to \mathcal{X}$. A program of length $K$ is an ordered sequence of primitives, $\tau = (z_{i_1}, \ldots, z_{i_K}) \in \mathcal{Z}^K$. Executing these primitives in order gives:
\begin{equation}
\label{eq:composition}
    f_\tau
    =
    f_{z_{i_K}}
    \circ
    f_{z_{i_{K-1}}}
    \circ
    \cdots
    \circ
    f_{z_{i_1}},
    \qquad
    y = f_\tau(x).
\end{equation}
In this formulation, a learner's \textbf{theory} of a phenomenon is represented as an executable program that explains the transformation from $x$ to $y$. The number of possible primitive sequences of length $K$ is $M^K$, so exhaustive coverage becomes increasingly costly as program length grows. This motivates learning reusable primitives that can be recombined into new executable theories to explain phenomena beyond those observed during training.

\textbf{Training data.}
L2T assumes a training dataset of i.i.d. observation pairs,
\begin{equation}
    \mathcal{D}_{\mathrm{train}}
    =
    \{(x_n,y_n)\}_{n=1}^{N},
    \qquad
    y_n = f_{\tau_n}(x_n),
    \qquad
    \tau_n \sim p_{\mathrm{train}}(\tau).
\end{equation}
Crucially, the programs $\{\tau_n\}$ and primitive execution functions $\{f_{z_i}\}$ are \textbf{latent and never observed}. The learner receives only pairs $(x_n,y_n)$, without program annotations, symbolic rules, action labels, or task groupings. Unlike few-shot abstract reasoning benchmarks such as ARC-AGI, where examples must be grouped by a shared hidden program, or imitation-learning and action-conditioned world-modeling settings, where transition-mediating actions are observed, L2T requires only raw observational pairs for training. For example, temporally separated trajectory observations, such as $(x_t, x_{t+t'})$, already form valid training data. Thus, L2T opens the possibility of large-scale training on uncurated observational datasets.

\textbf{Three capabilities.} Theorizing the world, in this formulation, requires a model to (i) \textbf{discover} reusable abstract primitives across phenomena, (ii) learn how to \textbf{compose} them into structured explanations, and (iii) \textbf{explain} novel phenomena by forming new compositions of the same primitives. In L2T, the model is evaluated at test time on programs drawn from a set $\mathcal{T}_{\mathrm{test}}$ disjoint from the programs used to generate training observations, including programs longer than those seen during training---jointly probing compositional and length generalization.

\textbf{Evaluating transferability.}
A central evaluation principle is \textbf{transferability}. Given two phenomena $\smash{(x^{(1)}, y^{(1)})}$ and $\smash{(x^{(2)}, y^{(2)})}$ generated by the same latent program $\tau$, a model first induces $\hat{\tau}$ from the first pair and then applies it to $x^{(2)}$ to predict $y^{(2)}$. Transferability measures whether $\hat{\tau}$ has captured a reusable mechanism rather than information specific to $\smash{(x^{(1)}, y^{(1)})}$. Self-explainability---re-applying $\hat{\tau}$ to $\smash{x^{(1)}}$---can be passed by memorization; transferability cannot.

\section{The BabyTheorist Environment}
BabyTheorist turns L2T into a controlled benchmark that tests whether learners can discover, compose, reuse, and update theories from observations alone. We make this setting concrete through three controls: (i) a program language defining hidden compositional programs (Section~\ref{sec:babytheorist_language}), (ii) frequent subprograms that recur across observations (Section~\ref{frequent_subprograms}), and (iii) a level-based curriculum that gradually increases program complexity (Section~\ref{sec:babytheorist_curriculum}). Building on these controls, BabyTheorist evaluates learners before and after training at each level (Section~\ref{sec:babytheorist_protocol}) and measures how far and how quickly they progress through the curriculum (Section~\ref{sec:babytheorist_metrics}).

\input{CameraReady/figures/fig_sec3_babytheorist_pipeline}

\subsection{The Program Language of BabyTheorist}
\label{sec:babytheorist_language}
\textbf{Observations.} The program language defines how BabyTheorist generates observations from hidden programs. Each observation pair consists of a before grid \(x\) and an after grid \(y\), generated by a hidden program that maps \(x\) to \(y\). Each grid is a \(15 \times 15\) array of colors. Color \(0\) denotes the background. Objects are connected colored shapes rendered with non-background colors \(1,\ldots,9\); they may vary in color pattern, position, and shape, as detailed in Appendix~\ref{app:objects_and_scenes}.

\textbf{Programs.} To apply a primitive to a scene, a program must specify which objects it acts on. Each program therefore consists of steps that pair a \emph{what} component, given by a selector, with a \emph{how} component, given by a primitive (Figure~\ref{fig:babytheorist_pipeline}a):
\[
\texttt{step(selector, primitive)}.
\]
The number of steps in a program is its \textbf{depth}, corresponding to the program length \(K\) from Section~\ref{phenomena_and_programs}.

\textbf{\textit{Selectors}} define the \emph{what}: the object or set of objects to which a step applies. BabyTheorist includes four kinds of selection rules: (i) global rules over all objects in the scene, (ii) attribute-based rules using object attributes such as color or hollowness, (iii) spatial rules based on positional criteria such as topmost or leftmost, and (iv) index-based rules using a fixed raster-scan ordering. The full selector vocabulary is given in Appendix~\ref{app:selectors}.

\textbf{\textit{Primitives}} define the \emph{how}: the type of change applied to the selected objects. BabyTheorist groups primitives into three categories: (i) transformation primitives, which modify an object's color, shape, interior, or boundary, (ii) movement primitives, which change an object's position, and (iii) relational primitives, which modify one object with respect to another. The full primitive vocabulary and execution semantics are given in Appendix~\ref{app:primitives}.

\input{CameraReady/tables/tab_sec4_bte_modes}
\subsection{Frequent Subprograms for Reusable Abstractions}
\label{frequent_subprograms}
Natural environments are not arbitrary collections of unrelated events. Some patterns recur across many situations and can appear within more complex patterns. BabyTheorist mirrors this structure by making some primitive sequences recur across different programs. We call these recurring sequences \textbf{frequent subprograms}. Reusing these subprograms can reduce the effective search space for explaining more complex observations. This design therefore rewards learners capable of discovering and compressing recurring structure into reusable abstractions.

\textbf{Subprogram pools.} To implement this structure, BabyTheorist maintains a frequent subprogram pool for each program depth, hidden from the learner (Figure~\ref{fig:babytheorist_pipeline}b). The depth-1 pool contains all available primitives. Deeper pools are built by combining shorter subprograms and keeping a limited number of the resulting sequences. Each entry preserves both its primitive sequence and which steps share a target. For example, ``move $\text{Object}_a$ up, then $\text{Object}_a$ right'' reuses one target, whereas ``move $\text{Object}_a$ up, then $\text{Object}_b$ right'' allows different targets. Targets are represented by placeholders, which receive concrete selectors during observation generation. The same subprogram can therefore recur across scenes while preserving its target relationships.

\textbf{Program sampling.} To generate a program of depth \(d\), the generator either takes a complete program from the depth-\(d\) pool or chains shorter pool entries, including single primitives, until their depths sum to \(d\); for example, the depth-6 program in Figure~\ref{fig:babytheorist_pipeline}b chains entries of depths 1, 3, and 2. For depths 2--10, each option is chosen with probability \(50\%\). As a result, each pool entry appears both as a program of its own depth and inside longer programs, so the same subprograms recur across observations of varying complexity. Appendix~\ref{app:frequent_subprograms} details pool construction, sampling, and primitive expansions.

\subsection{A Curriculum for Continual Theory System Updates}
\label{sec:babytheorist_curriculum}
A useful theory system should evolve by recognizing which phenomena its current theories can explain and which require a new theory. BabyTheorist studies this process through a \emph{level-based curriculum} with increasing program depth and an expanding primitive vocabulary. These changes require learners to build new theories from familiar primitives and to acquire new ones when needed.

\textbf{Per-level depth distributions.} We implement this curriculum by defining each level as a distribution over program depths rather than a single depth. As the curriculum advances, this distribution shifts toward longer programs while still including shorter ones. Together with the frequent subprograms introduced in Section~\ref{frequent_subprograms}, this exposes learners to structures that recur across levels and become components of more complex transformations. Appendix~\ref{app:depth_distribution} specifies the depth distribution.

\textbf{Keeping depth tied to complexity.} To make program depth informative about transformation complexity, we apply rule-based filtering to screen out \emph{degenerate cases}: programs whose observed transformations admit shorter explanations because of redundant or mutually cancelling steps (e.g., moving an object right and then left). Details of the rule-based filter are provided in Appendix~\ref{app_depth_filter}.

\textbf{Vocabulary expansion.} The curriculum can also introduce new primitives at later levels, adding one transformation primitive that alters an object's interior or boundary at each of Levels 2, 4, and 6 (Appendix~\ref{app:primitives}). The learner must then go beyond recombining learned primitives and expand its theory vocabulary. Each new primitive first appears within inherited subprograms and then joins the existing pools (Appendix~\ref{app:frequent_subprograms}).

\input{CameraReady/figures/fig_sec3_babytheorist_samples}

\textbf{Benchmark modes.} We define four benchmark modes (Table~\ref{tab:bte_modes}; examples in Figure~\ref{fig:babytheorist_samples}). The Single/Multi distinction controls scene structure: Single modes use one object, while Multi modes use multiple objects and relational primitives. The A/B distinction controls vocabulary expansion: A modes keep the primitive vocabulary fixed, while B modes introduce one new primitive at each of Levels 2, 4, and 6. Multi-B further increases the number of objects from two to three and expands the selector vocabulary to all selector types. As objects, selectors, and primitives increase, the program space grows rapidly, making theory-system expansion beneficial. Appendix~\ref{app:program_space_counts} gives the program-space counts.

\subsection{Evaluating Theory Reuse and Length Generalization}
\label{sec:babytheorist_protocol}
\begin{wrapfigure}{r}{0.3\textwidth}
  \vspace{-1.2em}
  \centering
  \includegraphics[width=\linewidth]{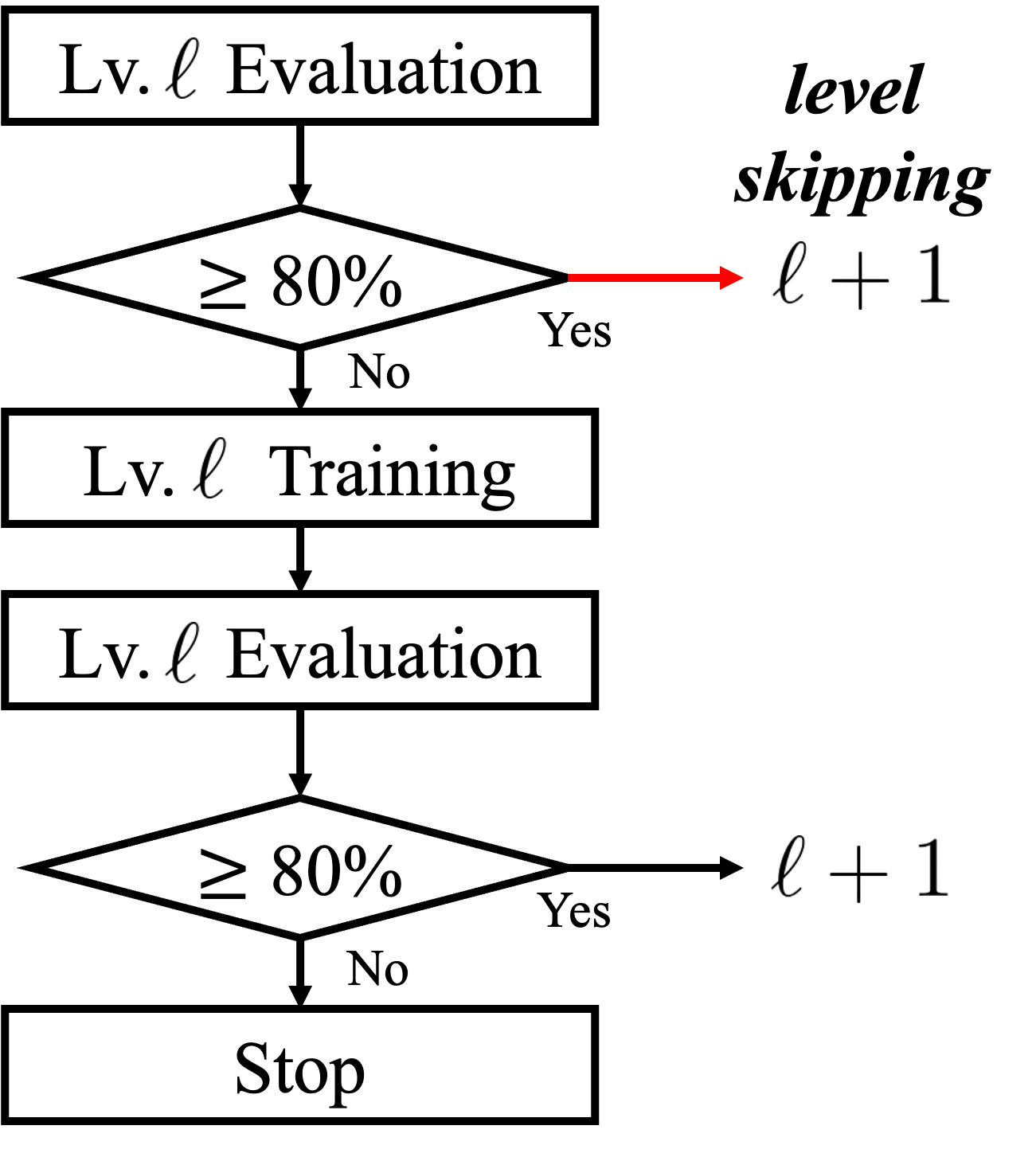}
  \caption{\textbf{Continual train--test protocol.} At the start of level \(\ell\), the learner is evaluated before training. Passing (at least \(80\%\) accuracy) enables \emph{level skipping} (red arrow): the learner advances without that level's training observations, rewarding transfer to increasingly complex levels.}
  \label{fig:continual_protocol}
  \vspace{-2.5em}
\end{wrapfigure}

To evaluate learners as the depth distribution shifts toward longer programs, BabyTheorist uses a \textbf{continual train-test protocol} over eight official levels (Figure~\ref{fig:continual_protocol}). At each level, the learner receives \(250{,}000\) training observations (Appendix~\ref{app:training_observations}) and then takes the level evaluation. It proceeds to the next level if it achieves at least \(80\%\) accuracy; otherwise, the curriculum ends. Notably, the protocol includes \textbf{level skipping}. Before training at each level, the learner takes an entry evaluation. If it meets the same passing threshold, it proceeds directly to the next level without receiving that level’s training observations. Successful skipping thus indicates length generalization to a more complex level, using the theories acquired earlier, without additional training on that level.

Each level evaluation consists of \(10{,}000\) tasks. Each task provides a support set of \(2\)--\(5\) before--after pairs generated by the same hidden program and a query input; the learner then predicts the query output. Importantly, the few-shot format is an evaluation device, not a training signal for theory acquisition. The supports reduce ambiguity about the task's hidden program. We start with two pairs and check whether the selection rules that explain them give the same query output. If not, we add more supports, up to five.

\subsection{Measuring Theory Learning Through Curriculum Progress}
\label{sec:babytheorist_metrics}
Because hidden programs, observation complexity, and shared subprogram structure are never labeled, our metrics are designed to reward discovering reusable theories rather than fitting provided rules or memorizing task identities. A strong learner should discover primitive theories, compose and reuse them, and expand the system as new primitives appear.

We rank learners by \textsc{Score} and use \textsc{Efficiency} to break ties. \textbf{\textsc{Score}}, our primary metric, measures how much of the curriculum the learner solves. Let \(a_{\ell}\) denote the exact-match accuracy of the evaluation that concludes level \(\ell\): the entry evaluation if skipped, otherwise the post-training evaluation. A task is solved only if the predicted output grid exactly matches the target. For example, solving \(8{,}000\) of the \(10{,}000\) tasks gives \(a_{\ell}=0.8\), earning the learner \(80\) points at that level. We sum these points as \(\textsc{Score}=\sum_{\ell=1}^{L_{\mathrm{reached}}}100\,a_{\ell}\), where \(L_{\mathrm{reached}}\) denotes the final curriculum level attempted by the learner.

\textbf{\textsc{Efficiency}} breaks ties in final \textsc{Score} by favoring learners that earn their points earlier in training. We count one \emph{training stage} for each level trained on. Skipped levels add points without increasing this count. Let \(S_t\) denote the cumulative \textsc{Score} after \(t\) training stages, including points from levels skipped before the next stage, and let \(S_{\mathrm{final}}\) be the final \textsc{Score}. The eight-level curriculum allows at most \(T_{\max}=8\) training stages. To compare learners over all \(T_{\max}\) stages, we set \(S_t=S_{\mathrm{final}}\) after a learner stops. \textsc{Efficiency} averages the fraction of final \textsc{Score} accumulated by each stage: \(\textsc{Efficiency}=\frac{1}{T_{\max}S_{\mathrm{final}}}\sum_{t=1}^{T_{\max}}S_t\).

Since BabyTheorist controls how observations are generated and how frequent subprograms are reused throughout the curriculum, high \textsc{Score} and \textsc{Efficiency} are intended to serve as behavioral evidence for observational theory learning. These objectives do not prescribe a particular theory representation: theories may be represented discretely, continuously, or implicitly within the learner.

\section{Related Works}

BabyTheorist relates to four lines of work. (i)~Program-induction and abstract-reasoning suites (ARC-AGI~\citep{chollet2019measure, arc24, arc25}, COGITAO~\citep{cogitao}, Bongard~\citep{nie2021bongardlogo, jiang2023bongardhoi}, RPM~\citep{zhang2019raven}) and (ii)~compositional-generalization suites (SCAN~\citep{scan}, COGS~\citep{cogs}, gSCAN~\citep{gscan}, CLEVR/CLEVRER~\citep{clevr, clevrer}) provide structured supervision, such as pre-grouped tasks or labeled primitive actions; BabyTheorist instead provides an \emph{ungrouped}, unlabeled observation stream in which hidden \emph{frequent subprograms} recur across a level curriculum.
 (iii)~Library-learning systems (DreamCoder~\citep{dreamcoder}, Bayesian program induction~\citep{lake2015human}) accumulate reusable abstractions on hand-curated suites, while BabyTheorist makes the abstraction set, its frequency, and its expansion controllable, and uses \emph{level skipping} to test reuse before a level's training observations are seen. (iv)~World modeling (PlaNet~\citep{planet}, Dreamer~\citep{dreamer, dreamerv2, dreamerv3}) and latent action learning (LAPO~\citep{lapo}, AdaWorld~\citep{adaworld}, LAPA~\citep{lapa}, Genie~\citep{genie}) primarily target rollout prediction or control; BabyTheorist instead evaluates whether explanations inferred from observations transfer to new inputs, rather than only reconstructing the observed transitions, as the hidden programs grow deeper across the curriculum.
 See Appendix~\ref{app:related_long} for full treatment.

\section{Experiments}

We evaluate whether existing methods acquire reusable theories from ungrouped observations. After reporting curriculum performance, we examine three challenges revealed by their predictions: discovering reusable primitives, learning where to apply a transformation, and explaining higher-depth phenomena. Each analysis connects observed failures to a research direction for observational theory learning.

\subsection{Baselines}

\label{sec:baselines}

We group baselines by how they represent the underlying program: \emph{implicit} models predict outputs without an explicit program latent, whereas \emph{explicit} latent models infer one, either as a single \emph{monolithic} latent or as a \emph{compositional} sequence of learned primitives.

\textbf{(i) Implicit models.} The \textbf{autoregressive Transformer (AR)} is trained on the same observation stream to generate the output grid token by token from the input, so any acquired theory remains implicit in its weights. We also include pretrained \textbf{LLMs} (GPT-OSS 120B and Qwen 3.5 27B), which are not trained on the stream and generate outputs in the same token-by-token manner.

\textbf{(ii) Monolithic latent models.} \textbf{Cont-Mono} and \textbf{Disc-Mono} infer a single continuous or discrete latent program from a before--after pair, respectively, following latent action models such as AdaWorld~\citep{adaworld} and LAPO~\citep{lapo}. A learned transition model then predicts the output from the input and this latent program. We consider two extensions of Cont-Mono. \textbf{Cont-Mono-Opt} refines the inferred latent program to fit the support pairs at test time, similar to LPN~\citep{lpn}. \textbf{Cont-Mono-OC} uses object-centric representations and infers a separate latent action for each slot, following FLAM~\citep{wang2026factoredlatentactionworld}.

\textbf{(iii) Compositional latent models.} \textbf{NEO}~\citep{baek2026l2t} infers a compositional program as an ordered sequence of discrete primitives from a shared vocabulary and executes their learned operations sequentially to predict outputs. This allows each learned primitive to be reused across many observed phenomena.

\textbf{Evaluation setup.} All models are evaluated on the same support--query tasks, but each uses the supports in a way suited to its representation, such as selecting among inferred programs, optimizing a latent, LoRA fine-tuning, or in-context prompting (Appendix~\ref{app:baselines_long}). Due to inference cost, we evaluate LLMs on the first 100 tasks at each attempted level. Architectures and hyperparameters appear in Appendices~\ref{sec:neo} and~\ref{app:hyperparam}.

\subsection{Overall Performance: Limited Progress Through the Curriculum}
\label{sec:exp_main}

\input{CameraReady/figures/fig_sec5_result}

\textbf{Every baseline stalls within the first two of the eight curriculum levels} (Figure~\ref{fig:main_results}; full results in Table~\ref{tab:main_results}, Appendix~\ref{app:main_results}). NEO alone reaches Level~2, only in the Single modes, and every learned baseline fails Level~1 in Multi. General-purpose reasoning, without learning the environment's rules from the observation stream, does not close this gap. The LLMs fail Level~1 in every mode, with GPT-OSS 120B and Qwen 3.5 27B reaching at most 59\% and 8\%, and even the frontier model GPT-5.6 Sol reaches only 51--64\% (Appendix~\ref{app:gpt56_sol}), well below the 80\% threshold. A diagnostic learner, PPO with Known DSL, helps locate the difficulty. Given the ground-truth primitives, selectors, and executor, it clears all eight levels of Single-A (Appendix~\ref{app:known_dsl_ppo}). It is not a matched comparison, since it never has to discover the operations, but it points to acquiring the operations as a major source of difficulty in the Single modes; selecting and composing them remains hard in richer multi-object settings, particularly Multi-B, even when they are provided. We next examine where the learned baselines fail, identifying three challenges for future methods.

\subsection{Challenge 1: Discovering Reusable Primitives}
\label{sec:challenge_primitives}

Before learners can build more complex theories, their basic transformations must work on new inputs. However, we found that models trained on the observation stream can explain all supports yet fail to transfer to the query, suggesting that their inferred programs fit the specific support pairs rather than capture a reusable transformation.

\begin{figure}[t]
\centering
\includegraphics[width=\linewidth]{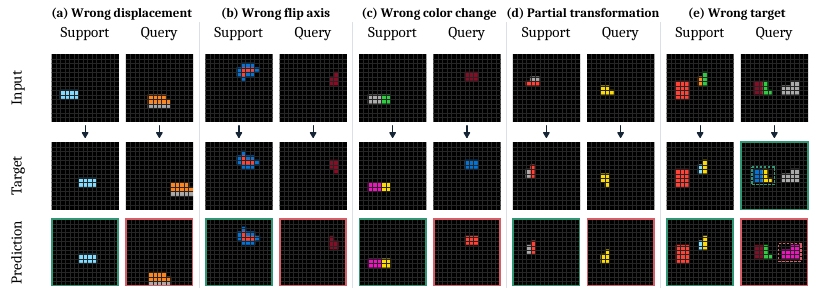}
\caption{\textbf{Transfer failures despite exact support fit.} Five failure patterns on Lv1 tasks: (a) incorrect displacement (AR), (b) incorrect flip axis (Cont-Mono), (c) two color shifts instead of one (Cont-Mono), (d) a rotation-only output when rotation followed by a flip is required (Disc-Mono), and (e) the correct color change applied to the wrong object (NEO; dashed boxes mark the changed object). Rows show Input, Target, and Prediction. Both supports are predicted correctly (one shown), but the query is incorrect; green/red borders mark correct/incorrect predictions. Examples (a--d) use Single-A/B and (e) uses Multi-A.}
\label{fig:baseline_failure_cases}
\vspace{-1em}
\end{figure}

\textbf{Explaining the Supports, Failing on the Query.} The gap appears even on single-object tasks requiring just one ground-truth primitive, without the additional demands of target selection or composition. Cont-Mono explains all supports yet fails to transfer to the query in 23.51\% of tasks, compared with 8.48\% for Disc-Mono and 2.36\% for NEO (Table~\ref{tab:support_transfer_seed}; examples in Figures~\ref{fig:baseline_failure_cases}a--c, \ref{fig:ar_support_transfer}, and~\ref{fig:single_action_transfer}; latent visualizations in Appendix~\ref{app:latent-visualization}). Cont-Mono-Opt refines the inferred latent within the learned program space to fit the supports, yet yields no consistent improvement in \textsc{Score} over Cont-Mono (Figure~\ref{fig:main_results}); in the depth-one analysis, the query errors it corrects are largely offset by new ones (Table~\ref{tab:failure_pattern_counts}; Figure~\ref{fig:opt10_regressions}). The results motivate learning a program space that favors reusable transformations in the first place. NEO's compositional representation, built from a shared primitive vocabulary, offers one design direction.

\subsection{Challenge 2: Learning Where to Apply a Transformation}
\label{sec:challenge_selection}

Reusing a primitive in a multi-object scene also requires knowing which object it should affect. We found that baselines with an explicit program latent, including NEO, can produce the correct primitive effect but apply it to the wrong query object, even after explaining all supports.

\textbf{The Right Transformation on the Wrong Object.} To examine target selection without multi-step composition, we analyze Multi-A Lv1, which uses index-based selectors throughout training and evaluation. Among depth-one tasks in which only one of multiple objects changes, NEO applies the correct transformation to a different object in 25.43\% of cases, averaged across three seeds (Table~\ref{tab:failure_pattern_counts}; Appendix~\ref{app:failure_patterns}). Figures~\ref{fig:baseline_failure_cases}e and~\ref{fig:target_transfer}c illustrate how such an error can arise. NEO changes the rightmost query object, although the task requires the first object in raster order. The two selection rules agree on both supports but select different objects in the query. The predictions therefore match a spatial rule that explains the supports but is absent from Multi-A's selector vocabulary. The example shows the need to learn the environment's selection conventions and apply them to new object layouts.

\textbf{Object-Centric Representations Still Need Selection Rules.} Making individual objects explicit is one way to support target selection. Cont-Mono-OC uses object slots and learns their alignment from the supports to transfer object-specific actions. However, separating objects does not by itself specify which one to change; that choice still requires a learned selection rule. Object-centric structure therefore does not automatically guarantee better transfer: Cont-Mono-OC's mean \textsc{Score} is 5.1 points higher than Cont-Mono's in Multi-B but 7.3 points lower in Multi-A (Table~\ref{tab:main_results}). Future work should investigate learning reusable selectors from the observation stream and composing them with primitive transformations, separating which object to change from how to change it.

\subsection{Challenge 3: Explaining Higher-Depth Phenomena}
\label{sec:exp_compositional}

\begin{figure}[t]
\centering
\includegraphics[width=\linewidth]{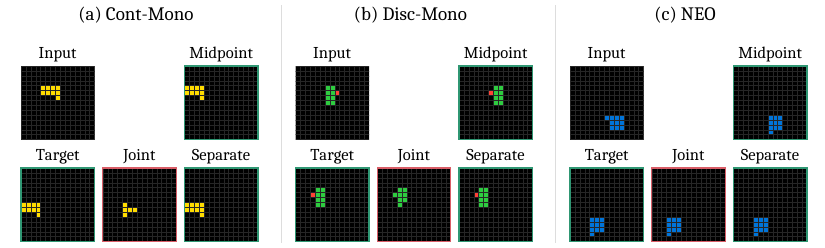}
\caption{\textbf{Step-wise inference succeeds where joint inference fails.} Joint predictions are wrong despite exact support fit. Given ground-truth intermediate supports, inferring a theory for each program step separately (\emph{Separate}) yields correct intermediate and final outputs. Each step's theory is inferred from its first support pair. This diagnostic adds supervision and changes inference, so it is not a baseline comparison.}
\label{fig:action_chaining_transfer}
\end{figure}

To explain higher-depth phenomena, learners must capture the combined effect of several primitive operations. Current methods struggle to meet this requirement in two ways. Models sometimes reproduce only part of the required change. Consistent execution of learned primitives remains a challenge even with compositional representations.

\textbf{Explaining the Whole Phenomenon.} We found cases where a model explains only part of the required change. In Figure~\ref{fig:baseline_failure_cases}d, Disc-Mono's output matches a rotation alone, although the phenomenon requires rotation followed by a flip. To probe this failure, we additionally give the model the ground-truth intermediate outputs of the supports and infer a theory for each step separately. Applied in sequence, these step-wise theories produce the correct query outputs for the cases in Figure~\ref{fig:action_chaining_transfer}, suggesting that the model can represent each step but struggles to decompose the combined change from before--after pairs alone. Moreover, on depth-two tasks, NEO achieves 88.2\% query accuracy, compared with 61.6\% for Disc-Mono (Table~\ref{tab:support_transfer_depth2}). This contrast motivates investigating compositional inductive biases that build complex explanations from reusable operations, as in NEO.

\textbf{Consistent Execution of Learned Primitives.} A learned primitive should have a consistent effect. However, we observed some cases where NEO's learned primitives produce inconsistent effects during execution (Appendix~\ref{app:state_dependent_execution}). NEO's state-grounding loss is intended to reduce such inconsistencies (Appendix~\ref{sec:neo}), but these cases motivate further study of its formulation and tuning.

\section{Limitations and Conclusions}
\label{sec:conclusion}

We introduced \textbf{BabyTheorist}, a controlled benchmark for studying \emph{observational theory learning}: acquiring reusable theories from streams of ungrouped before--after observations and applying them to new situations. The acquisition process is often left outside few-shot evaluations that present tasks already grouped by a hidden rule. Across seven representative baselines, all methods plateau well below ceiling. Compositional structure offers a promising direction, but even NEO---the strongest baseline---reaches only Level~2 in the Single modes and fails Level~1 in the Multi modes. Our analyses highlight difficulties in reusing primitives, selecting the intended objects, and explaining higher-depth phenomena. BabyTheorist is intentionally narrow---visually simple grids, a small primitive set, and deterministic dynamics---so the findings are limited to this controlled setting. With most of the curriculum still unsolved by the evaluated baselines, much remains to be done, and richer settings, real-world transfer, and the link to causal mechanism discovery are natural next steps.

\newpage
\section*{Acknowledgements}
This work was supported by the Institute of Information \& Communications Technology Planning \& Evaluation (IITP) grants funded by the Korea government (MSIT) (No. RS-2024-00509279, Global AI Frontier Lab; No. RS-2026-25518317, Development of AI memory mechanism that reflects human cognitive principles), the InnoCORE program of the Ministry of Science and ICT (AI Meta-Scientist, N10260110), and the Youlchon Foundation.

\section*{Broader Impact}
\label{app:broader_impact}
BabyTheorist is a benchmark for observational theory learning on synthetic grid-based observations; it contains no human, personally identifiable, or otherwise sensitive data, and its outputs are not directly deployable as user-facing systems. The intended positive impact is scientific: enabling controlled study of how learners acquire reusable, compositional theories from raw observations, which we believe is a prerequisite for sample-efficient, generalizable, and interpretable world models. Progress on this capability could in turn support downstream applications such as scientific discovery assistants, structured world models for embodied agents, and educational tools that probe and explain inferential reasoning. Because the benchmark is fully synthetic and procedurally generated, we see no immediate misuse vector beyond the general dual-use concerns that apply to any improvement in machine reasoning. We release the generator and evaluation code under the MIT License and the dataset under CC0~1.0 (links below the author list) to facilitate reproducibility and downstream research.

\bibliographystyle{unsrtnat}
\bibliography{CameraReady/refs_doojin,CameraReady/refs_ahn_,CameraReady/refs_ahn_local}

\newpage
\appendix
\input{CameraReady/99_appendix}

\ifincludechecklist
\clearpage
\input{checklist.tex}
\fi

\end{document}

%% file: CameraReady/figures/fig_sec1_observational_learning_general.tex
\begin{wrapfigure}{r}{0.34\textwidth}
  \centering
  \includegraphics[width=\linewidth]{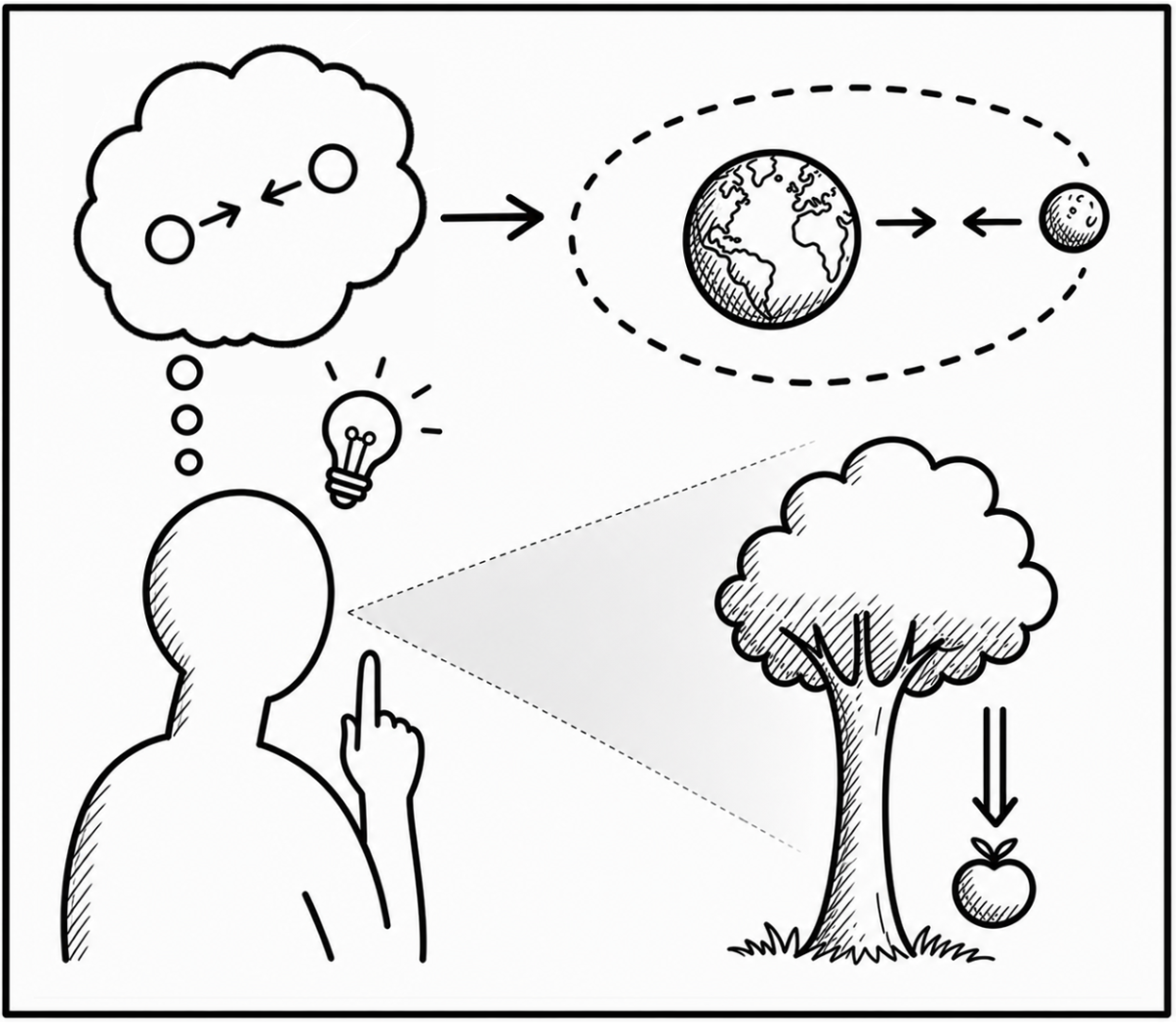}
  \caption{\textbf{Observational theory learning.} Humans learn about the world through theory acquisition and theory application. For example, observations of falling objects can lead to a theory of gravity, which can then be applied to explain orbital motion.}
  \label{fig:observational_learning}
  \vspace{-1.5em}
\end{wrapfigure}

%% file: CameraReady/figures/fig_sec1_overview.tex
\begin{figure}[t]
  \centering
  \includegraphics[width=\textwidth]{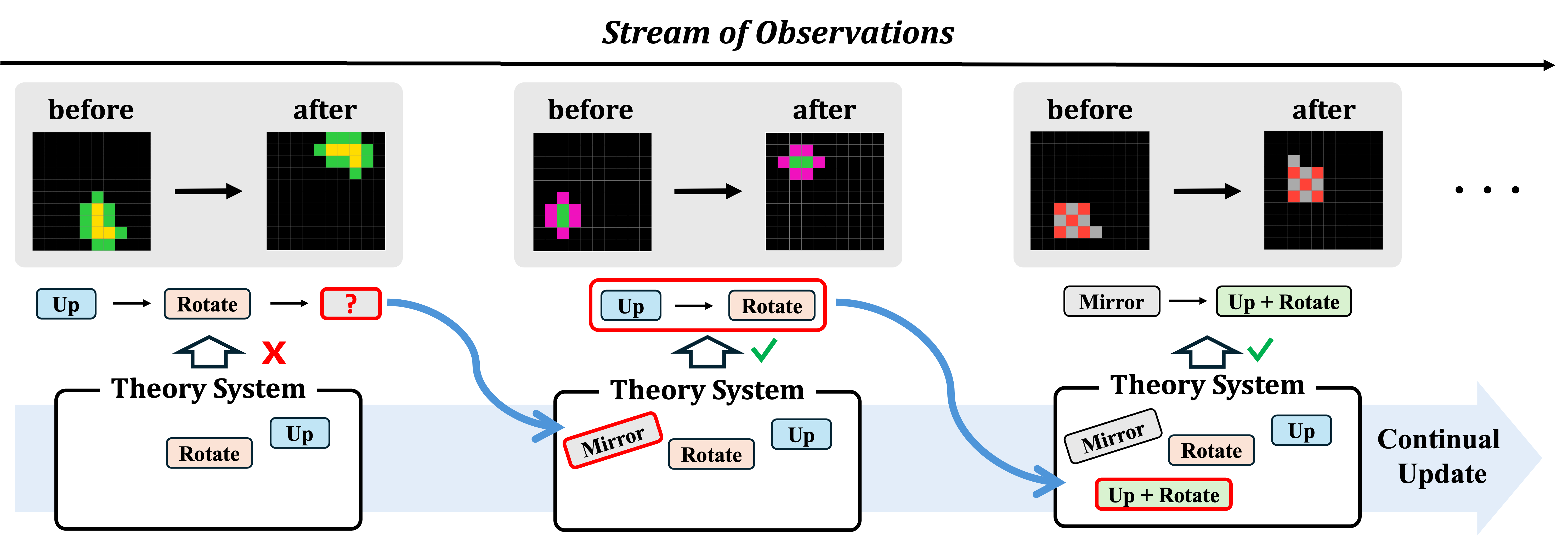}
  \caption{\textbf{Overview of continual observational theory learning.} A useful theory system should continually evolve as a learner receives a stream of observations. During \textit{theory acquisition} (blue arrows), the learner discovers new theories when existing ones cannot explain an observation (\textit{Mirror}) and abstracts recurring compositions into reusable units (\textit{Up + Rotate}). \textit{Theory application} combines them to explain new observations.}
  \label{fig:overview}
  \vspace{-1em}
\end{figure}

%% file: CameraReady/figures/fig_sec3_babytheorist_pipeline.tex
\begin{figure}[t]
  \centering
  \includegraphics[width=0.95\textwidth]{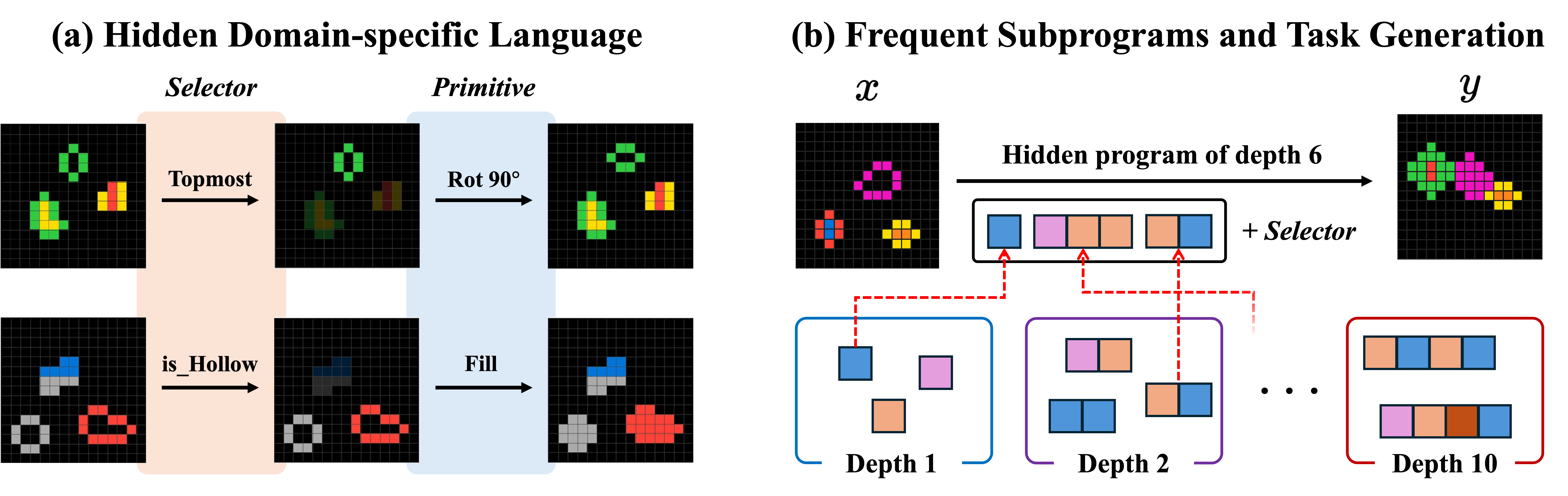}
  \caption{\textbf{BabyTheorist program language and task generation.} (a) Each program step pairs a selector, which identifies the objects to transform, with a primitive, which specifies the transformation. (b) The illustrated depth-6 program combines frequently recurring subprograms of depths 1, 3, and 2. Learners that discover such subprograms can reuse them to efficiently explain more complex phenomena.}
  \label{fig:babytheorist_pipeline}
  \vspace{-1em}
\end{figure}

%% file: CameraReady/tables/tab_sec4_bte_modes.tex
\begin{table}[t]
\centering
\small
\caption{\textbf{BabyTheorist benchmark modes.} The Single/Multi axis varies scene structure and relational reference, while the A/B axis varies whether the primitive vocabulary expands across levels. Program space counts nominal depth-\(d\) step sequences, including object/selector and primitive choices, after all primitives become available (Appendix~\ref{app:program_space_counts}).}
\label{tab:bte_modes}
\begin{tabular*}{1\textwidth}{@{\extracolsep{\fill}}lccccc@{}}
\toprule
Mode & \# Objects & Selectors & \# Primitives & Primitive Expansion & Program space \\
\midrule
Single-A & 1 & --  & 8 & -- & \(8^d\) \\
Single-B & 1 & --  & 11 & \checkmark (8 \(\rightarrow\) 11) & \(11^d\) \\
Multi-A & 2 & \texttt{by\_index} & 11 & -- & \(22^d\) \\
Multi-B & 3 & all selectors & 14 & \checkmark (11 \(\rightarrow\) 14) & \(216^d\) \\
\bottomrule
\end{tabular*}
\vspace{-1em}
\end{table}

%% file: CameraReady/figures/fig_sec3_babytheorist_samples.tex
\begin{figure}[t]
  \centering
  \includegraphics[width=\textwidth]{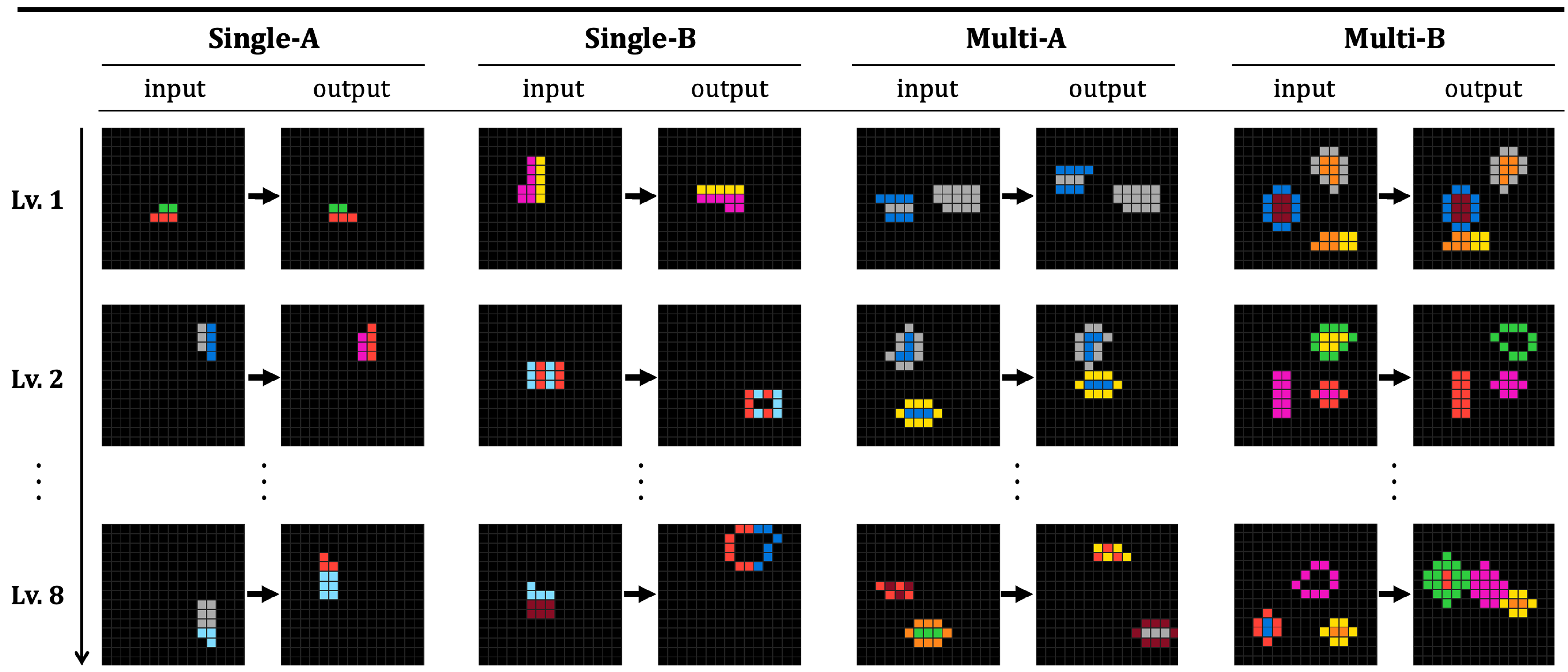}
  \caption{\textbf{Observations across modes and levels.} Each pair shows one before--after observation from Levels 1, 2, and 8 of each mode. Observations become more complex through longer hidden programs, more objects, relational structure, and primitive expansion.}
  \label{fig:babytheorist_samples}
  \vspace{-5mm}
\end{figure}

%% file: CameraReady/figures/fig_sec5_result.tex
\begin{figure*}[t]
\centering
\includegraphics[width=\textwidth]{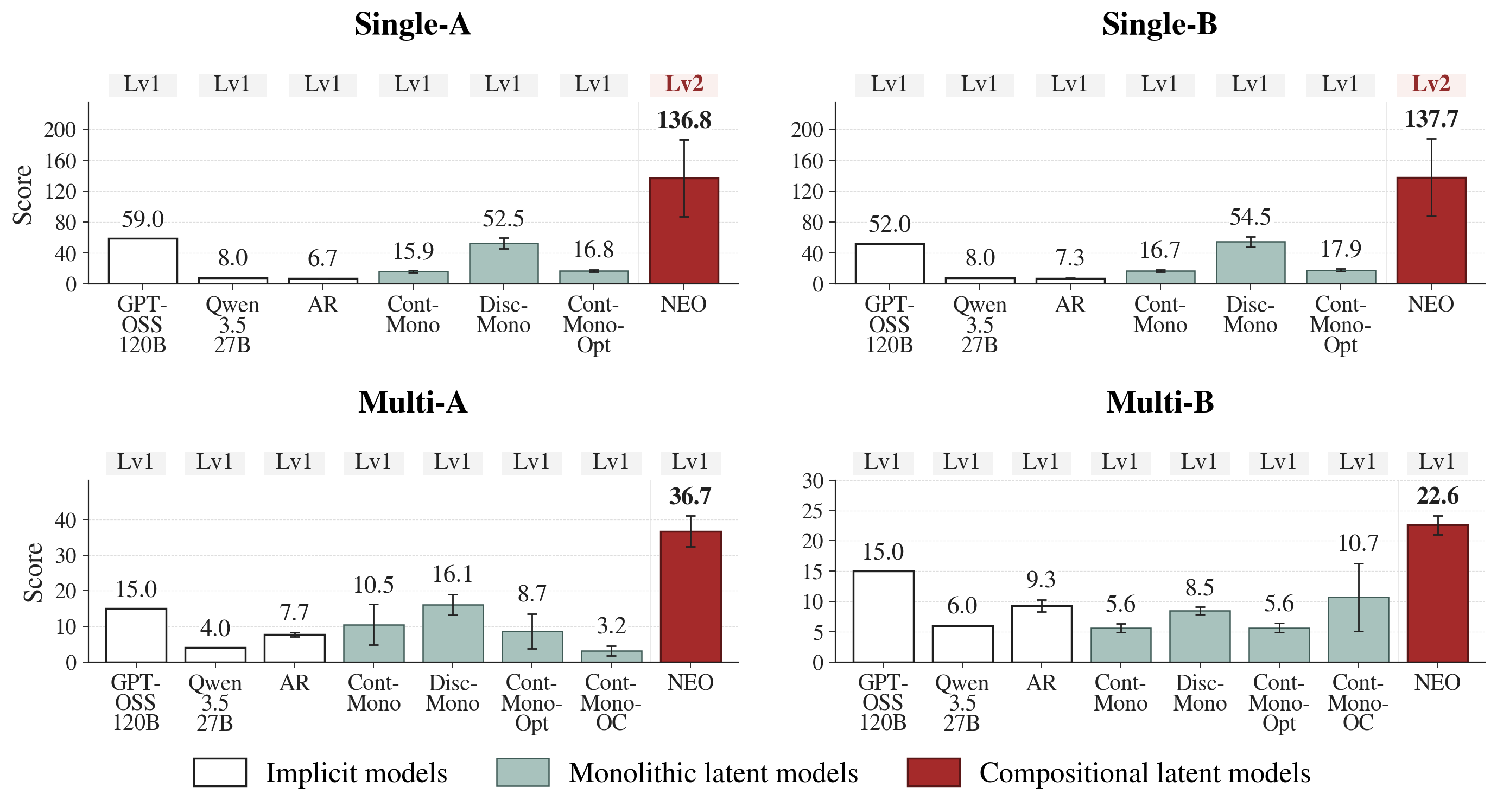}
\caption{\textbf{\textsc{Score} and highest curriculum level attempted in each benchmark mode.} No baseline progresses beyond Lv2. \textsc{Score} sums $100\times$ accuracy over attempted levels, so only models that reach Lv2 can exceed 100. Trained baselines show the mean $\pm$ one population standard deviation over three seeds; LLMs are evaluated once with greedy decoding on the first 100 tasks at each attempted level. Cont-Mono-OC is evaluated only on Multi-A and Multi-B. Labels above the bars give the highest level attempted.}
\label{fig:main_results}
\vspace{-5mm}
\end{figure*}

%% file: CameraReady/99_appendix.tex
\section{BabyTheorist Benchmark Details}
\label{app:program_language_figures}
\subsection{Objects and Scenes}
\label{app:objects_and_scenes}
\label{app:program_language_details}
Each BabyTheorist observation is generated on a \(15\times15\) grid with color \(0\) as background and colors \(1,\ldots,9\) as object colors. Objects are connected colored shapes. Their silhouettes are sampled from compact random walks, and their color patterns include solid, bordered, hollow, striped, split, and checkerboard variants. During scene generation, one or more such objects are placed on the grid subject to non-overlap and validity constraints. An object's position is the top-left corner of its bounding box, and its primary color is its most frequent color.

\subsection{Selectors}
\label{app:selectors}
\label{app:selector_vocabulary}
Selectors specify which object or object set a program step applies to. BabyTheorist uses eight selector forms: \texttt{ALL}, \texttt{by\_color(c)}, \texttt{is\_hollow}, \texttt{by\_index(k)}, \texttt{topmost}, \texttt{bottommost}, \texttt{leftmost}, and \texttt{rightmost}. Each selector picks its objects from the current scene at each step, not from the input scene.
The multi-select selectors are \texttt{ALL}, \texttt{by\_color(c)}, and \texttt{is\_hollow}. \texttt{ALL} selects every object in the scene; \texttt{by\_color(c)} selects objects whose primary color is \(c\); and \texttt{is\_hollow} selects objects with enclosed interior background pixels.

The single-select selectors are \texttt{by\_index(k)}, \texttt{topmost}, \texttt{bottommost}, \texttt{leftmost}, and \texttt{rightmost}. The \texttt{by\_index(k)} selector uses a 1-indexed raster ordering of objects in the input scene and remains tied to that object identity during execution. Positional selectors use deterministic tie-breaking: \texttt{topmost} chooses the smallest row, then smallest column; \texttt{bottommost} chooses the largest row, then largest column; \texttt{leftmost} chooses the smallest column, then smallest row; and \texttt{rightmost} chooses the largest column, then largest row.

\subsection{Primitives}
\label{app:primitives}
\label{app:primitive_vocabulary}
Primitives specify the change applied to selected objects. BabyTheorist contains fourteen primitives, grouped as follows.

\paragraph{Transformation primitives.}
The footprint-preserving transformations are \texttt{rot90}, \texttt{flip\_h}, \texttt{flip\_v}, and \texttt{next\_color}. They rotate, reflect, or recolor a selected object while keeping its position fixed. \texttt{rot90} rotates counter-clockwise, \texttt{flip\_h} flips upside down, \texttt{flip\_v} flips left to right, and \texttt{next\_color} cycles each color \(c\) to \((c \bmod 9)+1\). The footprint-changing transformations are \texttt{hollow}, \texttt{fill\_interior}, and \texttt{add\_border}. They remove interiors, fill interiors, or add a one-pixel outer border, respectively.

\paragraph{Movement primitives.}
The movement primitives are \texttt{translate\_up}, \texttt{translate\_down}, \texttt{translate\_left}, and \texttt{translate\_right}. Each moves the selected object by its own height or width in the corresponding direction.

\paragraph{Relational primitives.}
The relational primitives are \texttt{swap\_colors}, \texttt{align\_row}, and \texttt{align\_col}. Each takes a selected source object and a target object. \texttt{swap\_colors} swaps primary colors between the two objects; \texttt{align\_row} moves the source object to the target object's row; and \texttt{align\_col} moves the source object to the target object's column.

The base single-object vocabulary contains eight non-relational primitives: \texttt{rot90}, \texttt{flip\_h}, \texttt{flip\_v}, \texttt{next\_color}, \texttt{translate\_up}, \texttt{translate\_down}, \texttt{translate\_left}, and \texttt{translate\_right}. Multi-object modes add the three relational primitives. B modes additionally introduce \texttt{hollow}, \texttt{fill\_interior}, and \texttt{add\_border} at Levels 2, 4, and 6, respectively.

A program is valid only if every step changes the scene and keeps all objects inside the grid without overlap; a step with a multi-object selector applies to its objects one at a time.

\subsection{Nominal Program-Space Counts}
\label{app:program_space_counts}
Table~\ref{tab:bte_modes} reports a nominal depth-\(d\) program space for each benchmark mode after all primitives in that mode are available. These counts summarize the generator-side step-choice space before validity filtering, degeneracy filtering, and scene-specific selector failures.

In Single-A, there is one object and eight available primitives, so each step has \(8\) choices and a depth-\(d\) program has nominal space \(8^d\). In Single-B, the three additional object-shape primitives raise the maximum primitive count to \(11\), giving \(11^d\).

In Multi-A, there are two objects, the selector is \texttt{by\_index}, and the available vocabulary consists of eight non-relational primitives plus three relational primitives. A non-relational step has \(2\times8=16\) source-object/primitive choices. A relational step has \(2\) ordered source-target choices for each of the three relational primitives, giving \(3\times2=6\) choices. Thus each step has \(16+6=22\) nominal choices, yielding \(22^d\).

In Multi-B, the non-relational vocabulary has \(11\) primitives after all unlocks. The all-selector vocabulary gives \(18\) nominal selector instances over a 3-object scene: \(3\) \texttt{by\_index} choices, \(9\) \texttt{by\_color(c)} choices, four positional selectors, \texttt{ALL}, and \texttt{is\_hollow}. This gives \(18\times11=198\) non-relational choices. For relational primitives, the table counts concrete ordered source-target object choices, giving \(3\) source objects times \(2\) distinct targets times \(3\) relational primitives, or \(18\) choices. Therefore the nominal step count is \(198+18=216\), and the nominal depth-\(d\) space is \(216^d\). This is a compact benchmark-summary count; exact generation also samples selector parameters and applies symbolic and scene-level validity filters.

\subsection{Program Language Visual Examples}
\label{app:program_language_visual_examples}
\begin{figure}[!h]
\centering
\includegraphics[width=\textwidth]{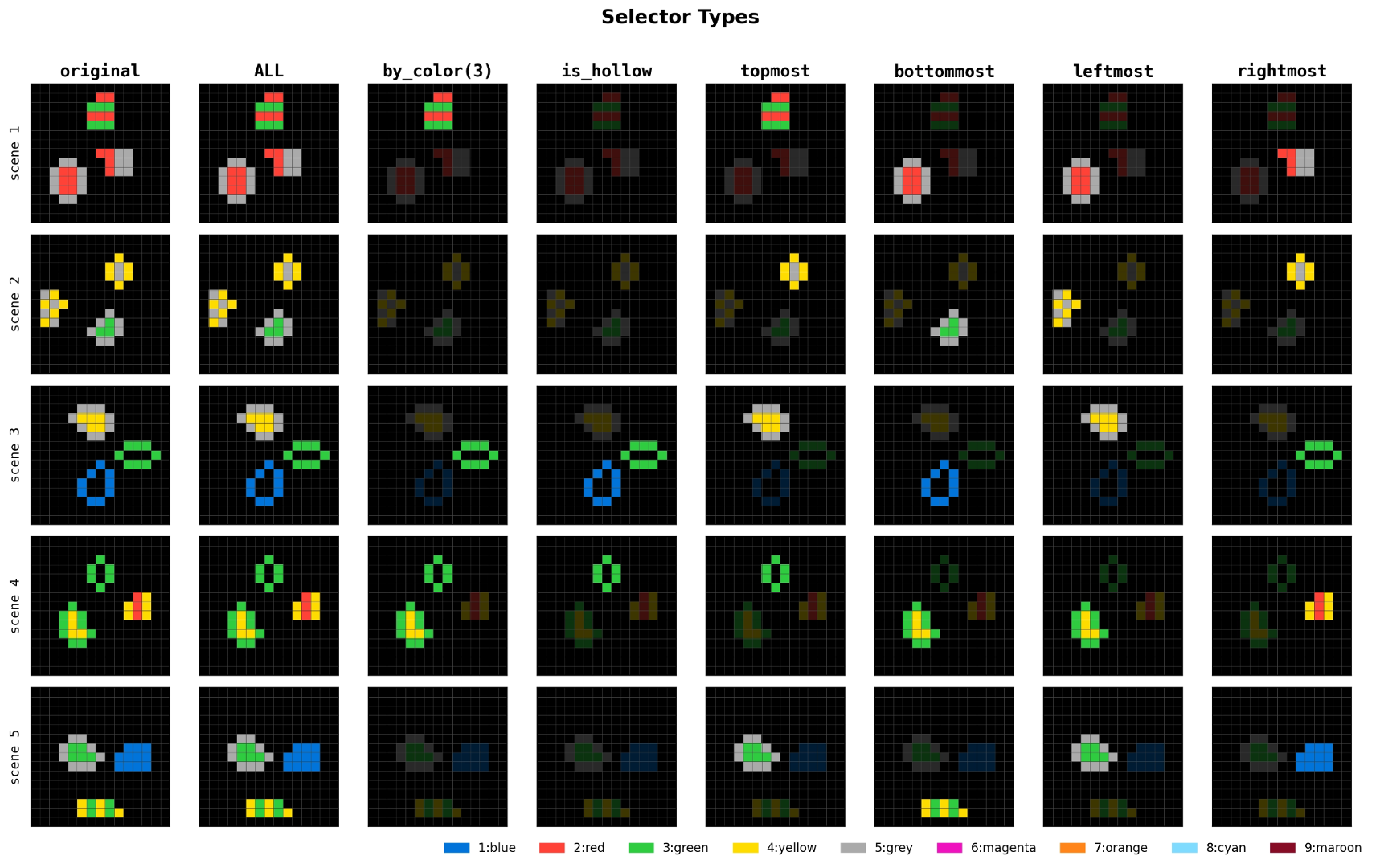}
\caption{\textbf{Selector types.} Each row shows a source scene and the objects selected by different selector types. Global selection applies to all objects; attribute-based selectors choose objects by properties such as color or hollowness; spatial selectors choose objects by position; and index-based selectors use a fixed raster-scan ordering.}
\label{fig:app_selectors}
\end{figure}

\begin{figure}[t]
\centering
\includegraphics[width=\textwidth]{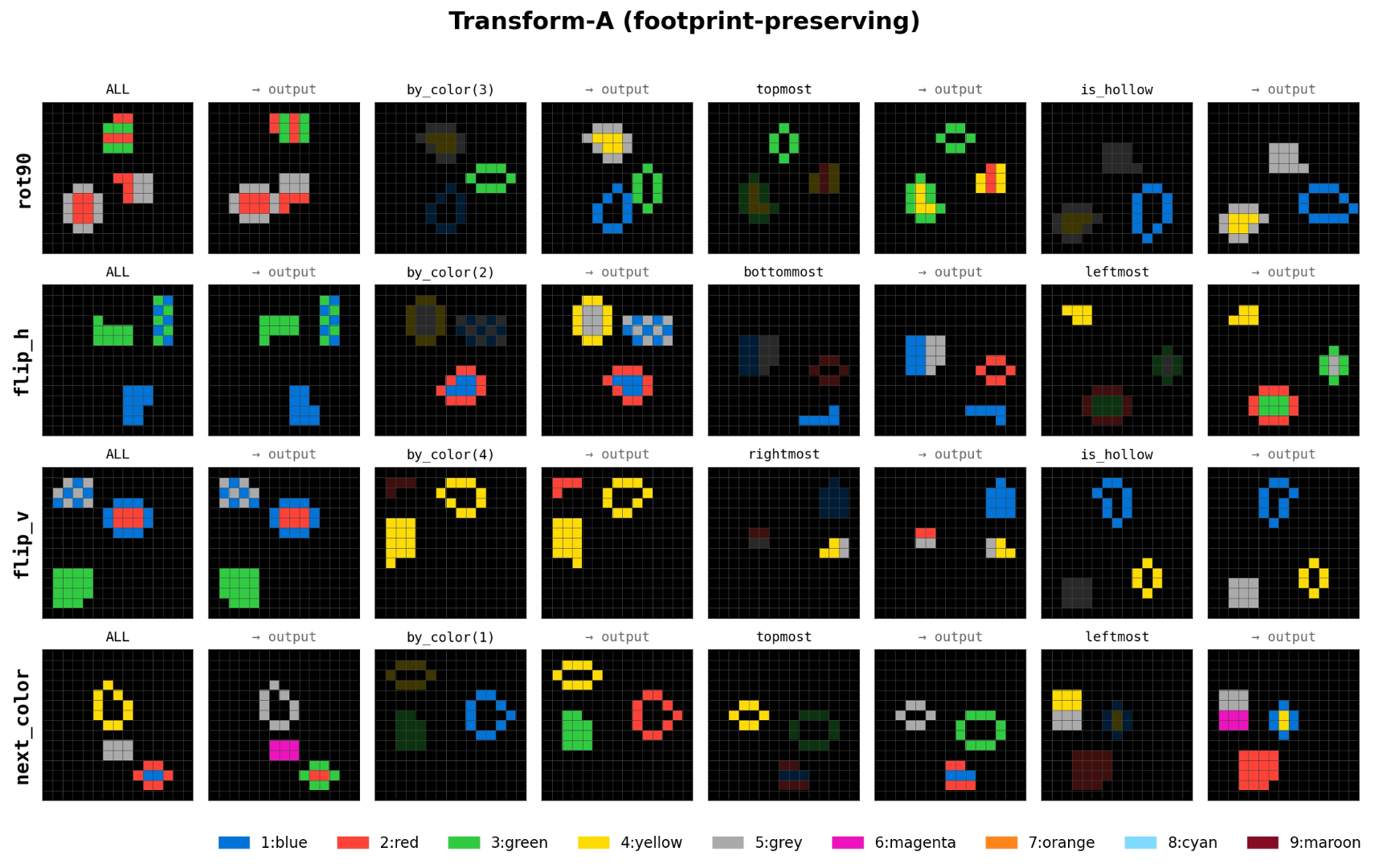}
\caption{\textbf{Footprint-preserving transformation primitives.} Representative before-after examples for primitives that transform selected objects while preserving their footprint, including rotation, reflection, and color change.}
\label{fig:app_primitives_transform_a}
\end{figure}

\begin{figure}[t]
\centering
\includegraphics[width=\textwidth]{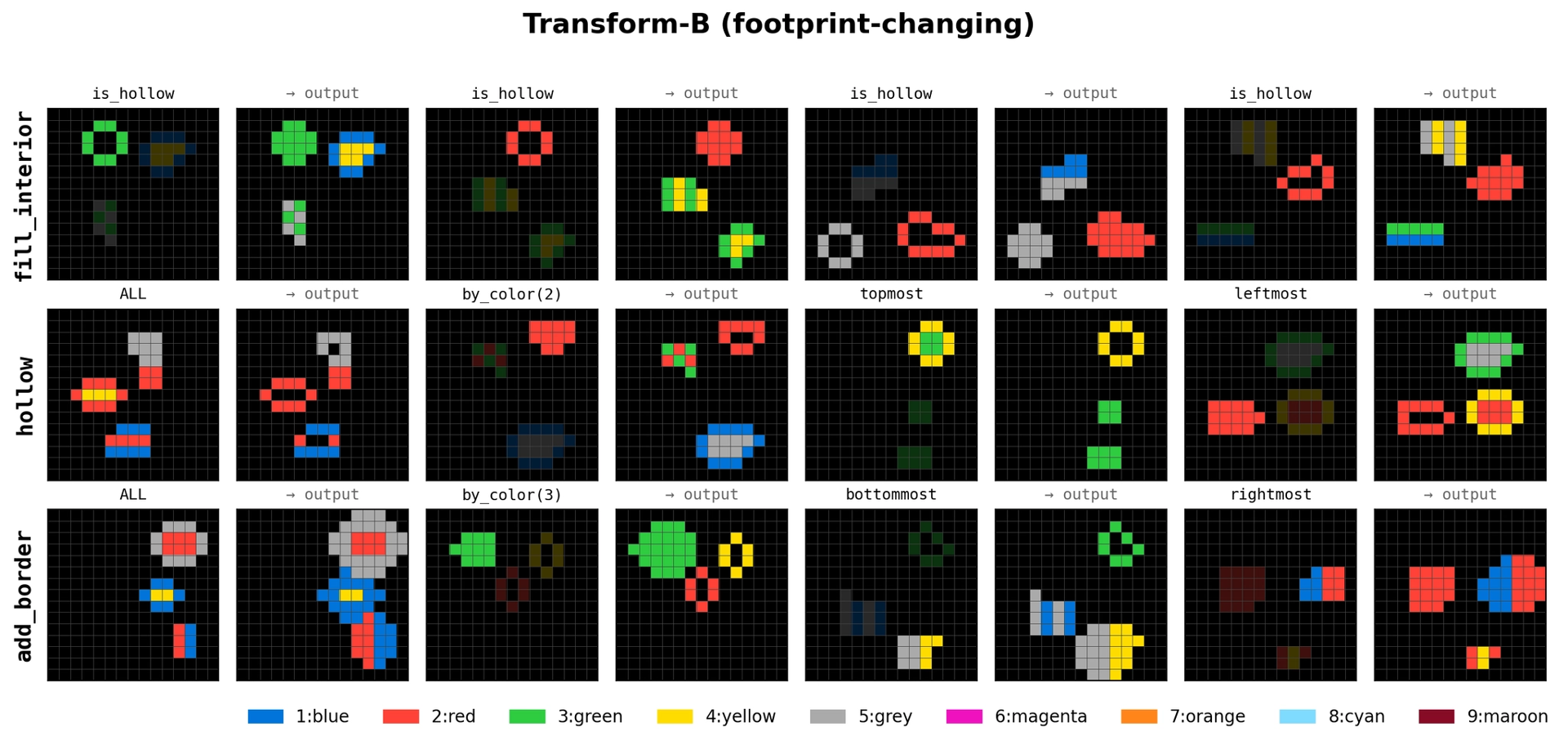}
\caption{\textbf{Footprint-changing transformation primitives.} Representative before-after examples for object-shape primitives that change the interior or boundary of selected objects, including \texttt{fill\_interior}, \texttt{hollow}, and \texttt{add\_border}.}
\label{fig:app_primitives_transform_b}
\end{figure}

\begin{figure}[t]
\centering
\includegraphics[width=\textwidth]{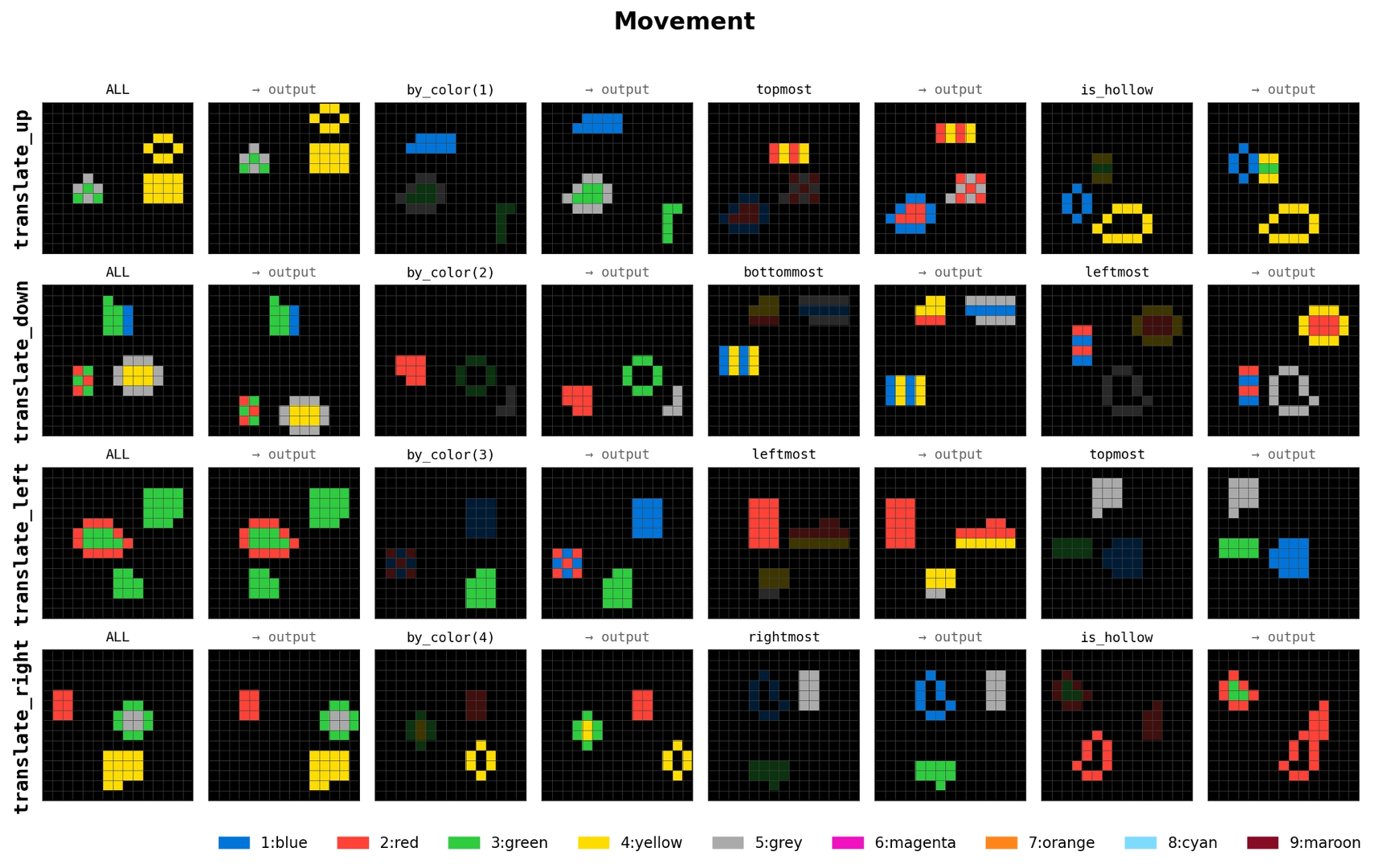}
\caption{\textbf{Movement primitives.} Representative before-after examples for primitives that translate selected objects up, down, left, or right.}
\label{fig:app_primitives_movement}
\end{figure}

\begin{figure}[t]
\centering
\includegraphics[width=\textwidth]{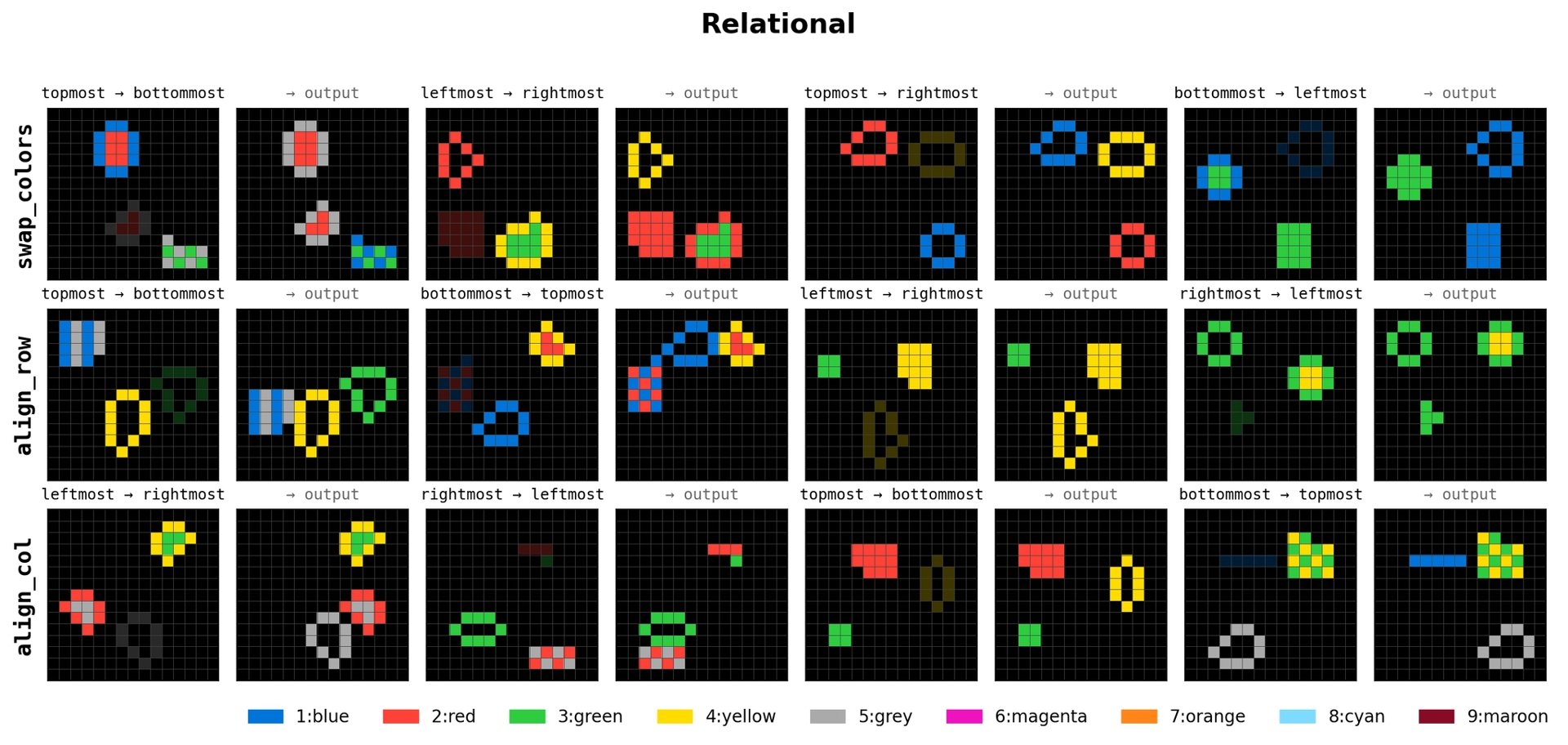}
\caption{\textbf{Relational primitives.} Representative before-after examples for primitives that use a target object, including color swapping and row or column alignment.}
\label{fig:app_primitives_relation}
\end{figure}

Figure~\ref{fig:app_selectors} visualizes the selector types used to bind a program step to an object or object set. Figures~\ref{fig:app_primitives_transform_a}--\ref{fig:app_primitives_relation} show representative primitive executions, grouped by whether the primitive preserves the object footprint, changes the object shape, moves objects, or uses a relational target.

\clearpage
\subsection{Frequent Subprogram Construction and Sampling}
\label{app:frequent_subprograms}
BabyTheorist maintains a frequent subprogram pool for each program depth. By sampling programs from these pools and combining their entries, the generator makes the same subprograms recur across programs of different depths. Below, we describe how primitive transition probabilities guide pool construction, how entries preserve their object relationships, and how the pools are used during program generation.

\textbf{Primitive transition probabilities.} To specify which primitives tend to occur together, the generator uses a transition matrix $T$, where $T_{ij}$ is the probability of choosing primitive $z_j$ after primitive $z_i$ in the primitive-sequence sampler. Each row sums to one. By default, the generator gives equal initial weight to all primitive pairs, sets known cancelling pairs to zero, and normalizes each row. For example, moving up followed immediately by moving down receives zero transition probability. Optional pair-specific weights can increase or decrease the relative probabilities of the remaining transitions.

These transition probabilities are used to score candidate primitive sequences. For a sequence $s=(z_{i_1},\ldots,z_{i_d})$, we compute
\begin{equation}
    w_{\mathrm{seq}}(s)
    = \left(\prod_{k=1}^{d} w_{i_k}\right)
      \left(\prod_{k=1}^{d-1} T_{i_k i_{k+1}}\right),
    \label{eq:frequent_sequence_weight}
\end{equation}
where $w_i$ is a configurable weight for primitive $z_i$, equal to one by default. Thus, the sequence weight accumulates the probabilities of its consecutive primitive pairs. For a concrete example under the default Single-A vocabulary, \texttt{translate\_up} has seven allowed successors, whereas \texttt{rot90} has eight. The sequence \texttt{translate\_up}, \texttt{rot90}, \texttt{next\_color} therefore has weight $(1/7)(1/8)=1/56$ before object bindings and filtering are considered. The weights belong to the data generator and are hidden from the learner.

\textbf{Representing the targets of operations.} A primitive sequence alone does not specify whether its steps act on the same object. For example, \emph{move up, then move right} can move one object twice or move two objects once each. Pool entries record this distinction using object placeholders: $A.\mathrm{up}\rightarrow A.\mathrm{right}$ shares one target, whereas $A.\mathrm{up}\rightarrow B.\mathrm{right}$ uses separate targets. Placeholders can also denote object groups; relational primitives record a source and a target. Concrete selectors and scene objects are assigned during observation generation. When entries are combined, their internal target relationships are preserved, and placeholders in different entries are renamed so that they are not accidentally tied together.

\textbf{Constructing the pools.} The depth-1 pool contains one valid entry for each available primitive. For greater depths, the generator combines shorter primitive sequences and samples candidates according to Equation~\ref{eq:frequent_sequence_weight}. It considers object-binding patterns that share one object, use two objects, share an object group, or use separate objects across steps. It also constructs candidates by combining entries from already validated, shallower pools. Candidates are checked using a temporary concrete assignment of their placeholders and the rule-based filter described in Appendix~\ref{app_depth_filter}.

To favor reusable target relationships, the generator applies a binding factor $b(\tau)=(1+0.25r)(1+0.10g)$, where $r$ counts repeated references to placeholders beyond their first occurrence and $g$ counts distinct group placeholders. A candidate obtained from a primitive sequence receives weight $w_{\mathrm{seq}}(s)b(\tau)$. A candidate formed by combining two validated entries receives the product of their stored weights multiplied by $b(\tau)$; duplicate candidates retain the largest weight assigned to them. Valid candidates are sampled without replacement in proportion to these weights. If too few survive filtering, additional primitive sequences are sampled using $T$ and checked in the same way. The base pools retain up to twenty entries per depth above one, with direct pools built through depth ten.

\textbf{Sampling programs from existing pools.} The generator first samples a program depth $d$ from the level-specific distribution in Appendix~\ref{app:depth_distribution}. For $2\leq d\leq10$, the basic pool sampler selects a complete entry from the depth-$d$ pool with probability $0.5$; otherwise, it combines shorter entries to reach depth $d$. Direct selection is proportional to the entries' stored weights. Depth-1 programs are sampled directly, while programs deeper than ten are always composed from pool entries.

For composition, the sampler repeatedly chooses a subprogram length uniformly from the allowable lengths and draws an entry from that pool. The first entry must be shorter than $d$, and subsequent entries cannot exceed the remaining depth or the maximum pool depth. When appending an entry, the sampler multiplies its stored weight by the transition probability between the last primitive already selected and the entry's first primitive, then normalizes the resulting weights. For example, a depth-8 program can combine entries of depths three, two, and three. An entry that appears in a shorter program can consequently reappear within a longer program.

\textbf{Adding new primitives in B modes.} Single-B and Multi-B preserve the existing pools when a new primitive is introduced. The generator constructs additional candidates by placing the new primitive between a prefix and a suffix drawn from inherited pools; either may be empty. Each candidate contains the new primitive exactly once and must pass the same filter before being added. The new pools retain up to ten entries per depth above one, weighted by the inherited entries' weights and the binding factor. Each new pool is sampled separately for two levels, then joins the inherited library.

The generator applies this two-level sampling window at Levels 2--3 for \texttt{hollow}, 4--5 for \texttt{fill\_interior}, and 6--7 for \texttt{add\_border}. During each window, it first decides whether to include the newly introduced primitive. For depth greater than one, it selects an entry from the newly added pools with probability $0.5$ and, if necessary, places inherited subprograms before and after it to reach the requested depth. Otherwise, it uses only inherited pools, following the basic sampler above. The depth of the new entry is selected uniformly among available depths not exceeding $d$, and the remaining depth is split uniformly between the prefix and suffix. This $0.5$ probability controls exposure to the new primitive; the basic sampler's $0.5$ probability controls direct selection versus composition within the inherited pools. Depth-1 sampling instead covers all currently available primitives.

\textbf{Generating observations.} After sampling an abstract program, the generator assigns concrete selectors and objects and applies the filter again, since concrete assignments can introduce additional interactions or redundancies. It then executes the program to generate before--after observations. Pool entries, transition probabilities, and generating programs remain hidden from the learner. All sampling probabilities above apply before filtering; the proportions among accepted observations can differ.

\subsection{Level Depth Distribution}
\label{app:curriculum_protocol}
\label{app:depth_distribution}
\begin{figure}[!h]
\centering
\includegraphics[width=\textwidth]{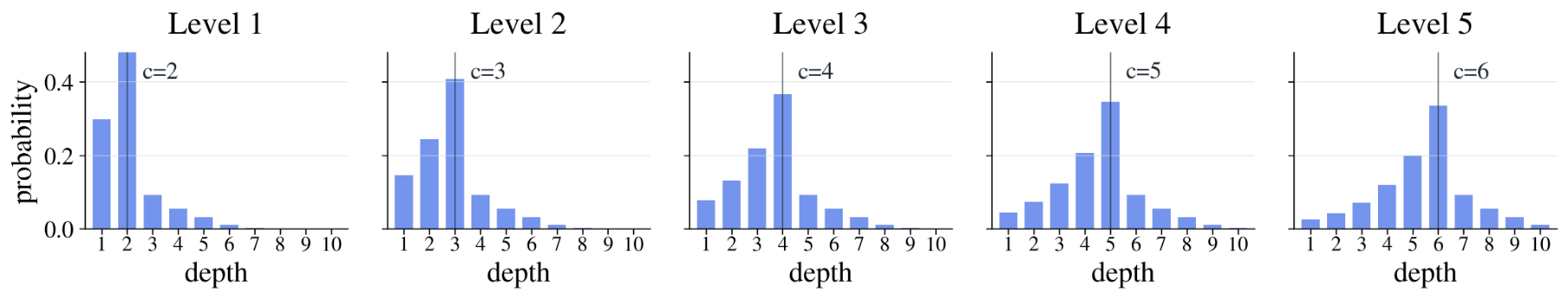}
\caption{\textbf{Level depth distributions.} Each level defines a distribution over hidden program depths rather than a single depth. As the level increases, the distribution shifts toward longer programs while still mixing observations of different complexity within the same level.}
\label{fig:app_depth_distribution}
\end{figure}

BabyTheorist defines each level as a distribution over program depths rather than a fixed depth. The implementation uses a simple three-part discrete prior. For level \(L\), let \(c=L+1\) be the center depth. The sampler allocates \(80\%\) probability mass to depths \(1,\ldots,c\), \(15\%\) to the shoulder depths \(c+1,c+2\), and \(5\%\) to a six-depth tail \(c+3,\ldots,c+8\):
\[
\begin{aligned}
p_L(d) &= 0.80\,q_{\mathrm{left}}(d)
       + 0.15\,q_{\mathrm{shoulder}}(d)
       + 0.05\,q_{\mathrm{tail}}(d),\\
q_{\mathrm{left}}(d) &\propto 0.60^{c-d}
       \quad &&\text{for } d\in\{1,\ldots,c\},\\
q_{\mathrm{shoulder}}(d) &\propto 0.60^{d-(c+1)}
       \quad &&\text{for } d\in\{c+1,c+2\},\\
q_{\mathrm{tail}}(d) &\propto 0.35^{d-(c+3)}
       \quad &&\text{for } d\in\{c+3,\ldots,c+8\}.
\end{aligned}
\]
Each \(q\) is normalized over its own support and is zero outside that support. Thus the full support is \(d\in\{1,\ldots,L+9\}\), and \(95\%\) of the mass lies at depths \(d\le c+2\). Figure~\ref{fig:app_depth_distribution} visualizes the resulting distributions across levels.

\subsection{Training Observations per Level}
\label{app:training_observations}
For each level not cleared by the entry evaluation, the learner receives 250,000 training observations. This number was selected based on preliminary experiments in which smaller datasets yielded limited learning progress. We keep it fixed across levels and trainable methods to standardize data availability.

\section{NEO: A Probabilistic Theory Induction Model}
\label{sec:neo}

\textbf{Generative model.}
NEO maximizes $p_\theta(y \mid x)$ by introducing two latent variables: a program $\tau = (z_{i_1}, \ldots, z_{i_K})$ and its execution trace $s = (s_1, \ldots, s_{K+1})$. Under a Markov assumption,
\begin{equation}
\label{eq:marginal}
p_\theta(y \mid x) \;=\; \int p_\theta(y \mid s_{K+1})\, p_\theta(\tau, s \mid x)\, d\tau\, ds,
\end{equation}
with the joint prior factorized as
\begin{equation}
\label{eq:prior}
p_\theta(\tau, s \mid x) \;=\; p_\theta(s_1 \mid x) \prod_{k=1}^{K} p_\theta(z_{i_k} \mid s_k) \, p_\theta(s_{k+1} \mid s_k, z_{i_k}).
\end{equation}
Three components are shared across this generative process: an \textbf{encoder} $E_\theta$ producing $s_1$ from $x$, a \textbf{theory programmer} selecting primitives, and a \textbf{shared executor} $f_\theta$ that realizes primitive semantics as latent state transitions.

\textbf{Training.}
Since the marginalization in Eq.~\eqref{eq:marginal} is intractable, NEO introduces a variational posterior
\begin{equation}
\label{eq:posterior}
q_\phi(\tau, s \mid x, y) \;=\; \underbrace{p_\theta(s_1 \mid x)}_{\text{encoder}} \;\prod_{k=1}^{K}\; \underbrace{q_\phi(z_{i_k} \mid s_k, y)}_{\text{theory programmer}} \;\underbrace{p_\theta(s_{k+1} \mid s_k, z_{i_k})}_{\text{execution}},
\end{equation}
and optimizes the resulting ELBO. The theory programmer is a \textbf{goal-conditioned policy}: at step $k$, it observes the current latent state $s_k$ and the encoded target $E_\theta(y)$, and selects the next primitive that steers the trace toward a state explaining $y$. Discrete primitive selection is implemented through a VQ-VAE codebook $\mathcal{E} = \{e_1, \ldots, e_{M'}\}$, where the codebook capacity $M'$ is a hyperparameter and need not match the (unknown) true number of primitives $M$.

\textbf{A learned Language of Thought.}
Primitives in NEO have no predefined meaning. Their \textbf{operational semantics} emerge from the shared executor: each codebook entry corresponds to a learned state transition $s_{k+1} = f_\theta(s_k, z_{i_k})$. NEO therefore jointly learns the \textbf{syntax} of programs (how primitives compose) and their \textbf{semantics} (what each primitive does), forming a learned Language of Thought directly from observation, without symbolic supervision.

\textbf{MDL and state grounding.}
Two design choices regulate learning. First, MDL-based length selection balances reconstruction error against program length,
\begin{equation}
\label{eq:mdl}
k^\star \;=\; \arg\min_{k \in \{1,\ldots,K+1\}} \;\lambda_{\mathrm{MDL}}^{\,k}\; \ell(y, \hat{y}_k), \qquad \lambda_{\mathrm{MDL}} > 1,
\end{equation}
penalizing length exponentially and biasing the model toward reusable primitives rather than overly fine-grained ones. Second, a \textbf{state-grounding} loss anchors each intermediate state $s_k$ to the manifold of valid encoder representations via a decode--encode consistency term, preventing shortcut transitions through arbitrary off-manifold states.

\textbf{Test-time inference.} Given $(x, y)$, NEO iteratively selects primitives from $q_\phi(z \mid s_k, y)$ and applies $f_\theta$ up to a configured maximum number of steps. It then selects the shortest prefix that exactly reconstructs $y$, or the prefix minimizing length-weighted reconstruction loss if no exact match is found. The same learned primitives can be composed repeatedly at inference time, within this step budget.

\newpage
\section{Full Baseline Details}
\label{app:baselines_long}
This appendix describes the architectures, training settings, and evaluation procedures of the baselines summarized in Section~\ref{sec:baselines}. All trainable baselines receive the same budget of 250,000 ungrouped observations per trained curriculum level. Evaluation uses support--query tasks and exact-match output-grid accuracy. Support usage is model-specific: LLMs place all support pairs in context, AR adapts LoRA parameters to the supports, and Cont-Mono-Opt optimizes a task-specific latent. Cont-Mono, Disc-Mono, and NEO use the Cross-Pair Validation procedure below with frozen model parameters. Cont-Mono-OC's evaluation procedure is described separately in its baseline paragraph. Comparisons therefore reflect both the program representation and the associated inference procedure.

\paragraph{Cross-Pair Validation.}
\label{app:cross_pair_validation}
For Cont-Mono, Disc-Mono, and NEO, we infer a candidate theory from each support pair using the trained inference model. Each candidate is applied to every support input, including the pair from which it was inferred, and its predictions are compared with the corresponding support outputs. We select the candidate most consistent with the full support set and apply it to the query input to predict the query output. All model parameters remain frozen throughout this procedure. The query output is used only to compute the final evaluation metric, not to infer or select a candidate. Unless otherwise specified, reported results for these three baselines use Cross-Pair Validation.

\paragraph{Large Language Model.}

We evaluate two open-weight LLM configurations: GPT OSS 120B with high reasoning effort and Qwen 3.5 27B with thinking disabled. This baseline tests whether general-purpose LLMs can solve BabyTheorist tasks through direct few-shot prompting alone, and contrasts with models that must acquire reusable theories from ungrouped observations before evaluation.
The full prompt is provided in Appendix~\ref{app:llm_prompt}.

\paragraph{Autoregressive Transformer (AR).}
A standard decoder-only Transformer~\citep{transformer} with RoPE positional encodings~\citep{su2023roformerenhancedtransformerrotary}, trained autoregressively under a causal mask on the ungrouped $(x, y)$ stream with no explicit program latent. Each $(x, y)$ pair is serialized into a single token sequence and the model is trained to predict $y$ tokens conditioned on $x$ tokens, optimizing the standard next-token log-likelihood. At evaluation, the frozen base model is LoRA-adapted~\citep{hu2021loralowrankadaptationlarge} on the supports for a fixed number of inner steps, with rank-8 adapters inserted into the attention query--key--value and output projections; we then generate $y_q$ autoregressively from $x_q$ under the adapted weights and discard the adapters before the next task. This baseline tests whether a generic sequence model with no inductive bias toward programs can, given a few supports, locate a parameter direction that recovers a transferable theory.

\paragraph{Continuous Monolithic (Cont-Mono).}
The theory inferred from a pair $(x, y)$ is represented by a single continuous vector $z \in \mathbb{R}^d$, with no internal compositional structure. An encoder $q_\phi(z \mid x, y)$ maps the pair to a Gaussian posterior over $z$, and a decoder $p_\theta(y \mid x, z)$ predicts $y$ from $x$ and a sampled $z$. The model is trained end-to-end as a conditional $\beta$-VAE~\citep{betavae}, with the $\beta$ term controlling the trade-off between reconstruction fidelity and posterior compactness. At evaluation, the posterior mean $\hat z_s = \mu_\phi(x_s, y_s)$ inferred from each support pair serves as a candidate continuous theory. We select among these candidates using Cross-Pair Validation (Appendix~\ref{app:cross_pair_validation}) and condition the decoder on the selected latent to predict $y_q$ from $x_q$. This baseline corresponds to continuous latent action models such as AdaWorld~\citep{adaworld}, and tests whether a single continuous theory vector suffices for transfer when no compositional bottleneck is imposed.

\paragraph{Discrete Monolithic (Disc-Mono).}
The theory inferred from a pair $(x, y)$ is represented by a single discrete vector $z \in \mathcal{E} = \{e_1, \ldots, e_{|\mathcal{E}|}\}$ drawn from a learned codebook. The encoder produces a continuous query that is quantized to its nearest codebook entry via straight-through estimation, and the decoder predicts $y$ from $x$ and $z$ as in Cont-Mono. The model is trained as a conditional VQ-VAE~\citep{vqvae}, with reconstruction, codebook, and commitment losses; the codebook size $|\mathcal{E}|$ is chosen per benchmark mode and held fixed. The theory is thus one symbol from a finite vocabulary, with no internal compositional structure. At evaluation, we infer a discrete theory candidate from each support pair and select the final candidate using Cross-Pair Validation (Appendix~\ref{app:cross_pair_validation}) before applying it to the query input. This baseline corresponds to discrete latent action models such as LAPO~\citep{lapo} and Genie~\citep{genie}, and tests whether discretization to a finite, non-compositional vocabulary already buys transfer over a continuous monolithic latent.

\paragraph{Continuous Monolithic with Latent Optimization (Cont-Mono-Opt).}
Cont-Mono-Opt reuses the trained Cont-Mono model and optimizes a task-specific latent at evaluation time. We initialize the latent by averaging the vectors inferred from all support pairs, then optimize it to fit the full support set while keeping all model parameters frozen. The optimized latent is applied to the query input.

\paragraph{Object-Centric Continuous Monolithic (Cont-Mono-OC).}
Cont-Mono with the monolithic input encoder replaced by Slot Attention~\citep{slotattention}, which decomposes each observation into $K$ slots, one of which is reserved for background, with a per-slot continuous latent $z_k \in \mathbb{R}^d$. A slot-level transition rolls each foreground slot forward independently conditioned on its own latent, and the predicted target is the aggregated reconstruction of the rolled-forward slots. The full model is trained end-to-end with the same conditional reconstruction objective as Cont-Mono, with no explicit slot-correspondence supervision; slot identity emerges from the iterative competition inside Slot Attention. Because slot order varies across images, we learn to align slots before transferring object-level latents from supports to the query. We fit a lightweight slot-ranking network per task, keeping the base model frozen. To train this network, we use the object-level latents inferred from one support pair to predict the outputs of the other supports and minimize the prediction error. This baseline corresponds to object-centric continuous latent action models such as FLAM~\citep{wang2026factoredlatentactionworld}, and tests whether object-centric factorization alone, without compositional primitives over time, suffices in BabyTheorist's Multi modes, where transformations apply to multiple objects simultaneously.

\paragraph{NEO.}
NEO~\citep{baek2026l2t}, described in Appendix~\ref{sec:neo}, represents a program as an ordered sequence of discrete primitives drawn from a learned codebook, with MDL-based length selection that penalizes longer programs. A goal-conditioned theory programmer selects one primitive at a time from the current latent state and the encoded target; the shared executor applies it, and the final state is decoded to predict $y$. At evaluation, we infer a candidate primitive sequence from each support pair and select the final sequence using Cross-Pair Validation (Appendix~\ref{app:cross_pair_validation}) before applying it to the query input. This baseline tests whether explicitly compositional discrete programs improve transfer over monolithic latents (continuous or discrete) in the absence of object-centric structure.

\subsection{LLM Prompt}
\label{app:llm_prompt}
The template below contains $K$ support input--output pairs and one query input. Each grid is serialized as a nested list of 15 rows, ordered from top to bottom, with 15 integer color values per row, ordered from left to right: 0 denotes background, and 1--9 denote object colors. Each grid placeholder is replaced with the complete $15\times15$ array. The ellipsis abbreviates the support list for presentation only; all support pairs are included in the actual prompt.

\begin{tcolorbox}[
  title={LLM Prompt},
  colback=white,
  colframe=black!75,
  colbacktitle=black!75,
  coltitle=white,
  fonttitle=\bfseries,
  fontupper=\small,
  boxrule=0.6pt,
  arc=2mm,
  left=3mm, right=3mm, top=3mm, bottom=3mm,
  toptitle=1.5mm, bottomtitle=1.5mm
]
\raggedright
\textbf{System message}\par\smallskip
You are a puzzle solving wizard.

\medskip
{\color{black!35}\hrule}
\medskip
\textbf{User message}\par\smallskip
Here are the example input and output pairs from which you should learn the underlying rule to later predict the output for the given test input:

\smallskip
\texttt{----------------------------------------}\par
{\color{blue!65!black}\ttfamily
\detokenize{[{'input': {support_input_1}, 'output': {support_output_1}},}\par
\detokenize{ ...,}\par
\detokenize{ {'input': {support_input_K}, 'output': {support_output_K}}]}\par
}
\texttt{----------------------------------------}
\smallskip

Now, solve the following puzzle based on its input grid by applying the rules you have learned from the training data.:

\smallskip
\texttt{----------------------------------------}\par
{\color{blue!65!black}\texttt{\detokenize{[{'input': {input_test_data}, 'output': [[]]}]}}}\par
\texttt{----------------------------------------}
\smallskip

What is the output grid? Only provide the output grid in the form as in the example input and output pairs. Do not provide reasoning or any additional information. Start with [[ and end with ]].:
\end{tcolorbox}

\newpage
\section{Extended Related Works}

\label{app:related_long}
We position BabyTheorist among existing benchmarks and ask, in turn, how each relates to \emph{observational theory learning}.

\paragraph{Abstract reasoning and program-induction benchmarks.}
A long line of benchmarks frames reasoning as program induction from a few demonstration pairs. ARC-AGI~\citep{chollet2019measure, arc24, arc25} is the canonical example, where a learner must recover the program behind a small support set and apply it to a held-out query. COGITAO~\citep{cogitao} extends this template into a generative grid-based visual reasoning framework that systematically samples compositions of primitive transformations, while Bongard-style tasks~\citep{nie2021bongardlogo, jiang2023bongardhoi} and Raven's Progressive Matrices~\citep{zhang2019raven} make the same demand visually. BabyTheorist shares the support--query evaluation format but differs in two design choices that target \emph{theory acquisition} rather than puzzle-solving. First, training data are \emph{ungrouped}: the learner sees a continual stream of input--output pairs with no task identities or support sets, and must discover that observations cluster around hidden programs. Second, beyond a controllable program language as in COGITAO, BabyTheorist also controls the distribution of recurring substructure and the complexity curriculum, enabling targeted study of how a theory system is built rather than how a fixed test set is solved.

\paragraph{Compositional generalization benchmarks.}
SCAN~\citep{scan}, COGS~\citep{cogs}, gSCAN~\citep{gscan}, and CLEVR/CLEVRER~\citep{clevr, clevrer} probe whether models trained on a finite set of primitive compositions can recombine them in novel ways. These benchmarks provide labeled task supervision and test recombination only at evaluation. BabyTheorist instead provides ungrouped observations and rewards models that recover the compositional structure itself, including \emph{frequent subprograms} that recur across observations as natural pressure for reusable theories.

\paragraph{Library learning and concept accumulation.}
Library-learning systems such as DreamCoder~\citep{dreamcoder} build up libraries of reusable abstractions over time by discovering recurring substructure in solved tasks; classical Bayesian program induction~\citep{lake2015human} similarly models concept acquisition as program learning over a generative grammar. These systems are typically evaluated on hand-curated domain suites where the abstraction set is implicit in the task design. BabyTheorist instead makes this abstraction set an explicit, controllable property of the data-generating process: \emph{frequent subprograms} are embedded in the observation stream, the curriculum expands the program vocabulary at known levels, and \emph{level skipping} directly measures whether previously acquired theories continue to explain new observations without further training.

\paragraph{World modeling and observation-only learning.}
Latent action models such as LAPO~\citep{lapo}, AdaWorld~\citep{adaworld}, LAPA~\citep{lapa}, and Genie~\citep{genie} infer latent actions from visual observations without ground-truth action labels. World models such as PlaNet~\citep{planet} and Dreamer~\citep{dreamer, dreamerv2, dreamerv3} learn action-conditioned dynamics from interaction data. Their evaluations target downstream control or rollout fidelity. BabyTheorist evaluates \emph{theory transfer}: whether explanations inferred from supports apply to held-out queries as the hidden programs grow deeper across the curriculum.

\newpage
\section{Filtering for Minimal Program Depth}
\label{app_depth_filter}

\paragraph{Depth can be misleading.}
BabyTheorist uses program depth as a controlled proxy for task complexity. However, naively sampling a program with \(d\) steps is not sufficient by itself. The rendered pair of input and output grids should not be explainable by another valid program with fewer than \(d\) steps. Otherwise, an example sampled at depth \(d\) may in fact behave like an easier example at depth 1 or 2. The generator therefore applies a symbolic depth filter before accepting a task.

\paragraph{Program based screening.}
The first part of this symbolic filter runs while the program is being sampled. It keeps a symbolic description of each object rather than trying all possible scenes. This description records (i) where the object would be, (ii) how it has been rotated or flipped, (iii) how its color has changed, and (iv) whether its shape has been bordered, hollowed, or filled. After each new operation, the filter checks whether the same effect could have been obtained with fewer steps. For example, it rejects programs where different operations cancel each other, or where several objects are changed one by one even though the same objects could have been selected as a group. These checks catch shortcuts that are already visible from the program structure, but they do not cover every possible shortcut.

\paragraph{Scene dependent shortcuts.}
The remaining shortcuts arise only after the concrete objects and their positions are chosen. For example, a rotation may be unnecessary on a symmetric object. An align operation can also behave like an ordinary move on a particular scene, which can then merge with other moves and make the whole program shorter. The second part of the filter handles these cases by building constraints on the input scene while the program is sampled. These constraints rule out object symmetries that would hide rotations or flips. They also avoid ambiguous shape sizes when rotations, translations, and borders are mixed. For color and position based selectors, the constraints force the sampled scene to realize the object set that the symbolic sampler intended. A task is accepted only after it passes both the program screen and the scene constraints, and the resulting transition changes the grid.

\paragraph{Oracle validation.}

To measure how reliable the filter is, we validate it with an independent oracle that is not used by the generator. For each task accepted by the symbolic filter, the oracle runs breadth first search (BFS) from the sampled input scene and tries every shorter program. It merges duplicate rendered states during the search, so it does not waste work on different program strings that lead to the same intermediate scene. If any shorter program reaches the same output grid, the task is counted as a shortcut.

\begin{table}[t]
\centering
\caption{Independent oracle validation of the depth filter on the Multi-B setting at level 4. We sampled 1,000 accepted tasks at each target depth and searched for shorter programs. Reliability is the fraction of accepted tasks for which the oracle found no shorter explanation.}
\label{tab_filter_bfs_audit}
\setlength{\tabcolsep}{6pt}
\begin{tabular}{lrrrr}
\toprule
Target depth & Samples & Validated & Shortcuts & Reliability \\
\midrule
1 & 1,000 & 1,000 & 0  & 100.0\% \\
2 & 1,000 & 998   & 2  & 99.8\% \\
3 & 1,000 & 997   & 3  & 99.7\% \\
4 & 1,000 & 987   & 13 & 98.7\% \\
\midrule
Total & 4,000 & 3,982 & 18 & 99.55\% \\
\bottomrule
\end{tabular}
\end{table}

Table~\ref{tab_filter_bfs_audit} shows that, on Multi-B tasks sampled at Level~4, the symbolic filter is highly reliable up to depth~4. Only \(18\) of \(4{,}000\) accepted samples had a shorter explanation found by the independent oracle. We therefore use the filter as a practical depth control mechanism, while treating depth as an empirically validated difficulty signal rather than a formal proof of minimal program length for every generated instance.

\newpage

\section{Model Hyperparameters}

\label{app:hyperparam}

All trainable baselines were run on two NVIDIA RTX 4090 GPUs (24 GB each), and each run completed within 24 hours of wall-clock time. GPT-OSS 120B inference was run on eight NVIDIA RTX PRO 6000 GPUs, and GPT-5.6 Sol was queried through the OpenAI API.

This appendix lists the architecture and training hyperparameters for every
model variant reported in the main paper. All shared backbone settings
(encoder / decoder / policy / transition) are identical across models; we
list them once per section and only flag the values that differ.

Common backbone (all variants):
\begin{itemize}
    \item Encoder / Decoder: Transformer with RoPE, $d_{\text{model}}{=}128$,
          $4$ layers, $4$ heads, $d_{\text{ff}}{=}512$, dropout $0$.
    \item Policy: MLP, $d_{\text{ff}}{=}512$, dropout $0$.
    \item Transition: AdaLN Transformer, $d_{\text{model}}{=}512$,
          $3$ layers, $4$ heads, $d_{\text{ff}}{=}2048$, dropout $0$.
    \item State: $8$ tokens of dim $512$.
    \item For models with explicit latent representations, each latent vector has $16$ dimensions. Cont-Mono, Cont-Mono-Opt, and Disc-Mono represent a program with a single vector, whereas NEO uses a sequence of vectors.
    \item Grid: $15{\times}15$, $10$ colours.
    \item Optimiser: AdamW, weight decay $0.01$, gradient clip $1.0$
          (except where noted), \texttt{bf16-mixed} precision.
\end{itemize}

Before any baseline-specific training, we pretrain the encoder / decoder
on the Lv1 training data of the corresponding mode. All subsequent
references to ``pretrained encoder / decoder'' refer to this checkpoint.

\subsection{AR (AutoRegressive)}
\label{app:hp-ar}

\begin{table}[htbp]
\centering
\caption{AR baseline hyperparameters.}
\label{tab:hp-ar}

\begin{tabular}{l|l}
\toprule
\textbf{Hyperparameter} & \textbf{Value} \\
\midrule
Backbone                 & Causal Transformer with RoPE \\
Hidden size              & $384$ \\
Number of blocks         & $12$ \\
Number of heads          & $8$ \\
Sequence length          & $451$ ($225 + 1 + 225$) \\
Vocabulary size          & $12$ \\
Prefix length            & $226$ ($225 + 1$ SEP) \\
Dropout                  & $0.1$ \\
Tie word embeddings      & false \\
Vocab lookup             & true \\
Loss on output tokens only & true \\
LoRA rank / alpha        & $8$ / $16$ \\
LoRA target modules      & attention QKV, attention output \\
LoRA dropout             & $0.0$ \\
Adaptation optimizer     & AdamW, LR $10^{-4}$, weight decay $0$ \\
Adaptation steps         & $400$ \\
\bottomrule
\end{tabular}
\end{table}

\subsection{Cont-Mono}
\label{app:hp-cont-mono}

Continuous monolithic latent regularized as a $\beta$-VAE, with Gaussian sampling during training and the posterior mean at evaluation, and no vector quantization. The observation encoder is deterministic and uses its posterior mean during both training and evaluation. All Cont-Mono runs are initialized from the pretrained encoder / decoder and trained end-to-end. Auto-reconstruction loss is enabled.

\begin{table}[htbp]
\centering
\caption{Cont-Mono training hyperparameters.}
\label{tab:hp-cont-mono}

\begin{tabular}{lccc}
\toprule
 & \textbf{Single Lv1} & \textbf{Multi-A Lv1} & \textbf{Multi-B Lv1} \\
\midrule
Seeds                       & 41, 42, 43            & 7, 13, 42             & 41, 42, 43 \\
Max transition length       & 1                     & 1                     & 1 \\
Codebook size               & ---                   & ---                   & --- \\
Length-control coeff        & 0.9                   & 0.9                   & 0.9 \\
Latent VAE $\beta$          & 0.05                  & 0.10                  & 0.05 \\
Recon loss weight           & 1.0                   & 1.0                   & 1.0 \\
Grounding loss weight       & ---                   & ---                   & --- \\
Auto-recon loss weight      & 5.0                   & 1.0                   & 5.0 \\
Batch size                  & 64                    & 64                    & 64 \\
Devices (DDP)               & 2                     & 2                     & 2 \\
Learning rate               & $5{\times}10^{-4}$    & $5{\times}10^{-4}$    & $5{\times}10^{-4}$ \\
LR schedule                 & constant              & constant              & constant \\
Warmup ratio                & 0.05                  & 0.05                  & 0.05 \\
Min LR ratio                & 0.005                 & 0.005                 & 0.005 \\
Epochs                      & 500                   & 500                   & 500 \\
\bottomrule
\end{tabular}
\end{table}

\subsection{Disc-Mono}
\label{app:hp-disc-mono}

Discrete monolithic latent with vector quantization (L2-normalised codebook, commitment cost $0.25$, no EMA). All Disc-Mono runs are initialized from the pretrained encoder / decoder and trained end-to-end.

\begin{table}[htbp]
\centering
\caption{Disc-Mono training hyperparameters.}
\label{tab:hp-disc-mono}

\begin{tabular}{lccc}
\toprule
 & \textbf{Single Lv1} & \textbf{Multi-A Lv1} & \textbf{Multi-B Lv1} \\
\midrule
Seeds                  & 41, 42, 43 & 42, 43, 44 & 42, 43, 44 \\
Max transition length  & 1 & 1 & 1 \\
Codebook size          & 40 & 80 & 320 \\
Recon loss weight      & 1.0 & 1.0 & 1.0 \\
Latent VQ loss weight  & 1.0 & 1.0 & 1.0 \\
Grounding loss weight  & 0.1 & 0.1 & 0.1 \\
Auto-recon loss weight & 5.0 & 5.0 & 5.0 \\
Batch size             & 128 & 128 & 128 \\
Devices                & 1 & 1 & 1 \\
Learning rate          & $5{\times}10^{-4}$ & $1{\times}10^{-3}$ & $1{\times}10^{-3}$ \\
LR schedule            & constant & cosine & cosine \\
Warmup ratio           & 0.05 & 0.005 & 0.005 \\
Min LR ratio           & 0.005 & 0.005 & 0.005 \\
Epochs                 & 500 & 500 & 500 \\
\bottomrule
\end{tabular}
\end{table}

\subsection{Cont-Mono-Opt (Test-time Latent Optimization)}
\label{app:hp-cont-mono-opt}

Cont-Mono-Opt does \emph{not} train any new model; it reuses the trained
Cont-Mono checkpoints (Section~\ref{app:hp-cont-mono}) and performs
test-time gradient search on the inferred latent $z$ with learning rate
$0.01$ for $10$ epochs.

\subsection{Cont-Mono-OC (Slot-based)}
\label{app:hp-cont-mono-oc}
Continuous object-centric action latents with a VAE regularizer and no vector quantization, with a separate action latent inferred for each slot. All Cont-Mono-OC runs are initialized from a pretrained object-centric autoencoder whose parameters are also updated during finetuning. Auto-reconstruction loss is enabled.
\begin{table}[htbp]
\centering
\caption{Cont-Mono-OC finetuning hyperparameters. Batch size is per GPU.}
\label{tab:hp-cont-mono-oc}
\begin{tabular}{lcc}
\toprule
 & \textbf{Multi-A Lv1} & \textbf{Multi-B Lv1} \\
\midrule
Seeds                       & 7, 13, 2024 & 7, 13, 2024 \\
Max transition length       & 1 & 1 \\
Codebook size               & --- & --- \\
Length-control coeff        & 0.1 & 0.1 \\
Latent VAE $\beta$           & 0.012 & 0.008 \\
Recon loss weight           & 1.0 & 1.0 \\
Grounding loss weight       & --- & --- \\
Auto-recon loss weight      & 1.0 & 1.0 \\
Number of slots             & 3 & 5 \\
Slot dimension              & 256 & 256 \\
Slot Attention iterations   & 5 & 5 \\
Action dimension per slot   & 16 & 16 \\
Batch size                  & 128 & 128 \\
Devices (DDP)               & 2 & 2 \\
Learning rate               & $7{\times}10^{-4}$ & $7{\times}10^{-4}$ \\
LR schedule                 & cosine & cosine \\
Warmup ratio                & 0.1 & 0.1 \\
Min LR ratio                & 0.1 & 0.1 \\
Epochs                      & 800 & 800 \\
\bottomrule
\end{tabular}
\end{table}
\subsection{NEO}
\label{app:hp-neo}

NEO is trained on top of the pretrained encoder / decoder. All NEO runs use vector quantization for the learned codes, with an L2-normalised codebook, commitment cost $0.25$, and auto-reconstruction loss.

\begin{table}[htbp]
\centering
\caption{NEO single-object hyperparameters.}
\label{tab:hp-neo-single}

\begin{tabularx}{\linewidth}{@{}>{\hsize=1.4\hsize\linewidth=\hsize\raggedright\arraybackslash}X*{4}{>{\hsize=.9\hsize\linewidth=\hsize\centering\arraybackslash}X}@{}}
\toprule
 & \multicolumn{2}{c}{\textbf{Lv1} (Single A \& B)} & \multicolumn{2}{c}{\textbf{Lv2}} \\
\cmidrule(lr){2-3} \cmidrule(lr){4-5}
 & Phase 1 (mtl=2) & Phase 2 (mtl=4) & Single-A & Single-B \\
\midrule
Seeds                  & 40, 41, 42 & 40, 41, 42 & 40, 42 & 40, 42 \\
Max transition length  & 2 & 4 & 4 & 4 \\
Codebook size          & 30 & 30 & 30 & 30 \\
Length-control coeff   & 0.9 & 1.0 & 1.2 & 1.2 \\
Recon loss weight      & 1.0 & 1.0 & 1.0 & 1.0 \\
Latent VQ loss weight  & 1.0 & 1.0 & 1.0 & 1.0 \\
Grounding loss weight  & 0.1 & 0.1 & 0.1 & 0.1 \\
Auto-recon loss weight & 5.0 & 5.0 & 5.0 & 5.0 \\
Batch size             & 64  & 64  & 128 & 128 \\
Devices (DDP)          & 2   & 2   & 1   & 1 \\
Learning rate          & $1{\times}10^{-3}$ & $1{\times}10^{-4}$ & $5{\times}10^{-5}$ & $1{\times}10^{-3}$ \\
LR schedule            & constant & constant & warmup\_\allowbreak{}constant & constant \\
Warmup ratio           & 0.0 & 0.05 & 0.05 & 0.05 \\
Min LR ratio           & 0.005 & 0.005 & 0.005 & 0.005 \\
Gradient clip          & 1.0 & 1.0 & 0.5 & 1.0 \\
Epochs                 & 500 & 200 & 50  & 50 \\
Init                   & pretrained enc / dec & Phase 1 ckpt & Phase 1 ckpt & Phase 1 ckpt \\
\bottomrule
\end{tabularx}
\end{table}

\begin{table}[htbp]
\centering
\caption{NEO multi-object hyperparameters.}
\label{tab:hp-neo-multi}

\begin{tabularx}{\linewidth}{@{}>{\hsize=1.4\hsize\linewidth=\hsize\raggedright\arraybackslash}X*{4}{>{\hsize=.9\hsize\linewidth=\hsize\centering\arraybackslash}X}@{}}
\toprule
 & \multicolumn{2}{c}{\textbf{Multi-A}} & \multicolumn{2}{c}{\textbf{Multi-B}} \\
\cmidrule(lr){2-3} \cmidrule(lr){4-5}
 & Phase 1 (mtl=2) & Phase 2 (mtl=3) & Phase 1 (mtl=2) & Phase 2 (mtl=3) \\
\midrule
Seeds                  & 42, 43, 44 & 42, 43, 44 & 42, 43, 44 & 42, 43, 44 \\
Max transition length  & 2 & 3 & 2 & 3 \\
Codebook size          & 40 & 40 & 160 & 160 \\
Length-control coeff   & 0.9 & 0.9 & 0.9 & 0.9 \\
Recon loss weight      & 1.0 & 1.0 & 1.0 & 1.0 \\
Latent VQ loss weight  & 1.0 & 1.0 & 1.0 & 1.0 \\
Grounding loss weight  & 0.1 & 0.1 & 0.1 & 0.1 \\
Auto-recon loss weight & 5.0 & 5.0 & 5.0 & 5.0 \\
Batch size             & 128 & 128 & 128 & 128 \\
Devices                & 1 & 1 & 1 & 1 \\
Learning rate          & $5{\times}10^{-4}$ & $5{\times}10^{-4}$ & $5{\times}10^{-4}$ & $5{\times}10^{-4}$ \\
LR schedule            & warmup\_\allowbreak{}constant & warmup\_\allowbreak{}constant & warmup\_\allowbreak{}constant & warmup\_\allowbreak{}constant \\
Warmup ratio           & 0.005 & 0.005 & 0.005 & 0.005 \\
Min LR ratio           & 0.005 & 0.005 & 0.005 & 0.005 \\
Gradient clip          & 1.0 & 1.0 & 1.0 & 1.0 \\
Epochs                 & 500 & 500 & 500 & 500 \\
Init                   & pretrained enc / dec & Phase 1 ckpt & pretrained enc / dec & Phase 1 ckpt \\
\bottomrule
\end{tabularx}
\end{table}

\newpage

\clearpage
\section{Additional Results}
\label{app:additional_results}

\subsection{Full Score Results}
\label{app:main_results}
Table~\ref{tab:main_results} lists the \textsc{Score} values shown in Figure~\ref{fig:main_results}. Trained baselines report the mean and population standard deviation over three seeds. LLMs are evaluated once with greedy decoding on the first 100 tasks at each attempted level, so they have no standard deviation. Cont-Mono-OC is evaluated only in the Multi modes.

\input{CameraReady/tables/tab_sec5_result}

\subsection{PPO with Known DSL}
\label{app:known_dsl_ppo}

To assess how difficult program inference and composition remain when the operations are already available, we evaluate \textbf{PPO with Known DSL}. This learner receives the ground-truth primitive and selector vocabulary available at each level, together with the exact executor. It must still infer a program that explains the observations, but it does not need to discover the operations or learn their execution semantics. We therefore report it separately from the learning-focused baselines in Section~\ref{sec:baselines} as a diagnostic with privileged DSL information.

\subsubsection{Learning to Select Program Steps}

Training uses independent before--after pairs $(x,y)$, with no ground-truth programs, program depths, or task groupings. For each pair, an episode starts from $x$ and builds a program one step at a time. The policy observes the current grid and the target $y$, selects a DSL action, and receives the resulting grid from the exact executor. A Single-mode action selects a primitive for the sole object. In Multi modes, the policy selects a source selector, a primitive, and, for relational operations, a target selector. These decisions specify one executable step. For execution, objects are recovered from eight-connected components of non-background pixels in the input grid; their identities are retained across subsequent steps.

The policy is Markov: each decision depends on the current grid, target grid, and available DSL vocabulary, without a recurrent state or previous-action input. A three-layer convolutional encoder processes the current and target grids together with grid-validity and coordinate channels. Its features condition the action distribution and value estimate. The Single-mode policy scores complete actions directly; the Multi-mode policy factorizes an action into the conditional source, primitive, and target choices described above. Proximal policy optimization (PPO)~\citep{schulman2017proximal} trains the policy and value function from the resulting trajectories.

To provide feedback before an entire transformation is solved, the reward combines exact reconstruction with progress toward the target. Let $x_t$ be the current grid, with $x_0=x$, and let $D(u,v)$ count mismatched pixels. We define
\begin{equation}
\label{eq:known_dsl_reward}
\begin{aligned}
\Phi(z) &= \operatorname{clip}\!\left(\frac{D(x,y)-D(z,y)}{D(x,y)},-1,1\right), \\
r_t &= \mathbf{1}\{x_{t+1}=y\}+\Phi(x_{t+1})-\Phi(x_t).
\end{aligned}
\end{equation}
The potential $\Phi$ measures progress relative to the initial mismatch, while the indicator gives a bonus for exact success. Training pairs have $x\ne y$. An action that cannot be executed leaves the state unchanged, receives zero reward, and consumes one step. An episode ends on exact success or after 17 steps. The learner receives neither a state-dependent mask identifying executable actions nor auxiliary search-generated action labels.

Each level requiring training supplies $250{,}000$ observations. We collect one trajectory per observation in a single pass through that level's data and perform four PPO optimization epochs per rollout batch. The policy carries over to subsequent levels. Table~\ref{tab:known_dsl_hyperparameters} gives the configuration used for all four modes and seeds 0, 1, and 2.

\begin{table}[htbp]
\centering
\small
\caption{Training and inference settings for PPO with Known DSL. The policy head differs between Single and Multi modes as described in the text.}
\label{tab:known_dsl_hyperparameters}
\begin{tabular}{@{}ll@{}}
\toprule
Setting & Value \\
\midrule
CNN channels / hidden width & $(64,128,128)$ / $512$ \\
Action-component embedding / action feature width & $64$ / $256$ \\
Optimizer / learning rate & Adam / $1.5\times10^{-4}$ \\
Adam $\epsilon$ / gradient-norm clipping & $10^{-5}$ / $0.5$ \\
Rollout batch / episode minibatch size & $512$ / $128$ \\
PPO epochs / clipping coefficient & $4$ / $0.2$ \\
Entropy / value-loss coefficient & $0.1$ / $0.5$ \\
Discount $\gamma$ / GAE $\lambda$ & $1.0$ / $0.95$ \\
Advantage normalization & Within each minibatch \\
Pixel-progress coefficient & $1.0$ \\
Beam width / branch width / maximum steps & $8$ / $16$ / $17$ \\
\bottomrule
\end{tabular}
\end{table}

\subsubsection{Inferring a Program from the Support Set}

At evaluation, the learned policy guides a beam search for one program that reproduces every support output. Each partial program is executed on all support inputs. To rank its possible next actions, we average the policy's log probabilities over supports whose current grids do not yet match their targets; these scores accumulate along the program. We consider actions in score order, retain up to 16 continuations per partial program that execute successfully on every support, and keep the eight highest-scoring partial programs for the next step. Search is limited to 17 steps.

A completed candidate must reproduce all support outputs exactly. Among candidates that also execute successfully on the query input, the planner selects the highest-scoring program, preferring a shorter program when scores tie. Executing that program produces the query prediction. If search finds no such program, the prediction defaults to the unchanged query input. The query output is never used during program search or selection, and the policy parameters remain fixed throughout evaluation.

\subsubsection{Results and Interpretation}

We evaluate on BabyTheorist v0.5.5, using all $10{,}000$ evaluation tasks at each attempted level. Entry evaluations permit level skipping; a failed entry evaluation triggers training on that level's $250{,}000$ observations. A level is passed at $80\%$ exact-match accuracy; the curriculum stops at the first post-training failure. Table~\ref{tab:known_dsl_results} reports mean \textsc{Score} and population standard deviation across three seeds.\footnote{The archived runner allowed repeated post-training submissions. Across all 12 runs, these repetitions used identical checkpoints and predictions. Retaining only the first post-training evaluation leaves the reported \textsc{Score}s and level progression unchanged.}

\begin{table}[htbp]
\centering
\caption{PPO with Known DSL on the four BabyTheorist modes. \textsc{Score} is reported as mean $\pm$ population standard deviation over three seeds. Level progression is identical across seeds; zero passed levels means failure at Level 1. Settings were selected using the public evaluation set, as described below.}
\label{tab:known_dsl_results}
\begin{tabular}{@{}lccc@{}}
\toprule
Mode & \textsc{Score} & Highest level passed & First failed level \\
\midrule
Single-A & $680.2\pm2.5$ & 8 & -- \\
Single-B & $500.4\pm0.8$ & 5 & 6 \\
Multi-A & $331.8\pm1.5$ & 3 & 4 \\
Multi-B & $52.5\pm5.9$ & 0 & 1 \\
\bottomrule
\end{tabular}
\end{table}

Training settings were selected using public Multi-B Level-1 evaluations, and branch width 16 was selected after comparing widths 8 and 16 on the public evaluation set. These results therefore describe a diagnostic configuration selected with evaluation feedback. They should not be interpreted as performance on an untouched test set or as a matched comparison with learners that must acquire the DSL from observations.

With the operations and executor provided, this planner completes Single-A and progresses through five levels of Single-B and three of Multi-A. It still fails to pass Level 1 of Multi-B in every seed. The remaining difficulty involves proposing and combining actions under a finite search budget, even when their execution semantics are known. These results characterize the combined learned policy and search procedure: before PPO training, the same search procedure already achieves $84.8$--$85.5\%$ accuracy at Level 1 of both Single modes. They do not establish an upper bound on performance with a known DSL.

\subsection{GPT-5.6 Sol: Performance and Error Analysis}
\label{app:gpt56_sol}

We additionally evaluate GPT-5.6 Sol to examine how well a pretrained reasoning model can infer transformations from support examples. The model serves as a general-purpose reference and receives no training on BabyTheorist's observation stream. Using high reasoning effort, we evaluate the first 100 tasks from each of the eight levels in all four modes of BabyTheorist v0.5.5, totaling 3,200 tasks. Each task is evaluated independently without tools, using the support--query prompt in Appendix~\ref{app:llm_prompt}. Predictions are scored by exact match with the target grid.

GPT-5.6 Sol achieves 51--64\% accuracy at Level 1, below the 80\% passing threshold in every mode. This result cannot be explained by output-format errors alone: even counting every malformed Level-1 prediction as correct would leave accuracy at or below 69\% in all four modes. To examine performance on more complex tasks, we also evaluate Levels 2--8 independently; these evaluations do not count as curriculum progression. As shown in Table~\ref{tab:gpt56_sol_levels}, accuracy generally decreases at later levels, reaching 30--40\% at Level 8.

\begin{table}[htbp]
\centering
\small
\setlength{\tabcolsep}{4.5pt}
\caption{GPT-5.6 Sol exact-match accuracy (\%) on 100 tasks per mode and level. Levels 2--8 are evaluated independently for diagnostic purposes.}
\label{tab:gpt56_sol_levels}
\begin{tabular}{@{}lrrrrrrrr@{}}
\toprule
Mode & Level 1 & Level 2 & Level 3 & Level 4 & Level 5 & Level 6 & Level 7 & Level 8 \\
\midrule
Single-A & 58 & 68 & 58 & 55 & 47 & 44 & 37 & 31 \\
Single-B & 64 & 69 & 60 & 57 & 51 & 39 & 45 & 40 \\
Multi-A & 51 & 48 & 42 & 50 & 42 & 42 & 31 & 30 \\
Multi-B & 52 & 48 & 51 & 53 & 40 & 29 & 35 & 33 \\
\bottomrule
\end{tabular}
\end{table}

To characterize the errors already present at Level 1, we compare its 400 predictions with the target grids and intermediate outputs of the generating programs. These programs are used only for post-hoc analysis and are not provided to the model. Of the 175 incorrect predictions, 21 have malformed grids, leaving 154 parseable but incorrect outputs.

\paragraph{More steps, more mistakes.}
Level 1 includes both single-step and multi-step programs. Accuracy decreases from 67.8\% on single-step programs (103/152) to 49.2\% on longer programs (122/248). This gap appears in all four modes and remains after excluding malformed outputs. Some predictions also reproduce only part of the required transformation. For example, one Multi-A task requires a left--right reflection of one object followed by two $90^\circ$ rotations of another. The prediction exactly matches the grid after the first step, leaving the second object's required rotation absent from the output.

\paragraph{The right objects in the wrong places.}
In 89 of the 154 parseable errors (57.8\%), the prediction can be reproduced exactly by changing only the positions of the target objects, preserving their shapes and internal colors without adding or omitting foreground pixels. Forty of these cases also reproduce required appearance changes, so they cannot be explained by simply relocating unchanged input objects. The remaining error lies in where the objects are placed. For example, a two-cell-high object that should move upward by its own height is instead moved upward by three cells. Its appearance and movement direction are correct, but the displacement is not.

\paragraph{Relations pose a greater challenge.}
In Multi-B, accuracy is 31.8\% on tasks containing relational operations such as alignment or color swapping (14/44), compared with 67.9\% on the remaining tasks (38/56). Lower accuracy on relational tasks is also observed within the depth-1 and depth-2 subsets. Multi-A shows a different pattern, with corresponding accuracies of 61.5\% and 44.3\%. Thus, this analysis identifies relational tasks as a difficult subset of the sampled Multi-B setting, rather than establishing a general disadvantage for relational operations.

These comparisons reveal failures to reproduce complete transformations and exact spatial placement. They characterize the predicted outputs, but do not determine whether the model inferred an incorrect rule or failed to execute an intended rule accurately.

\setlength{\intextsep}{6pt}\setlength{\textfloatsep}{8pt}\setlength{\floatsep}{6pt}
\section{Failure Cases}
\label{app:failure_analysis}\label{app:case_summary}
This appendix gives the protocols, statistics, and examples supporting the three challenges in Sections~\ref{sec:challenge_primitives}--\ref{sec:exp_compositional}. We first define the diagnostic scope, then present primitive transfer and latent refinement, target selection, and compositional execution. Each labeled panel group shows one task. S1/S2 denote supports and Q the query; green/red borders indicate correct/incorrect outputs. All grids retain their original coordinates.

\subsection{Evaluation scope and support--query statistics}
\label{app:failure_scope}
\begin{figure}[!ht]
\centering
\includegraphics[width=\linewidth]{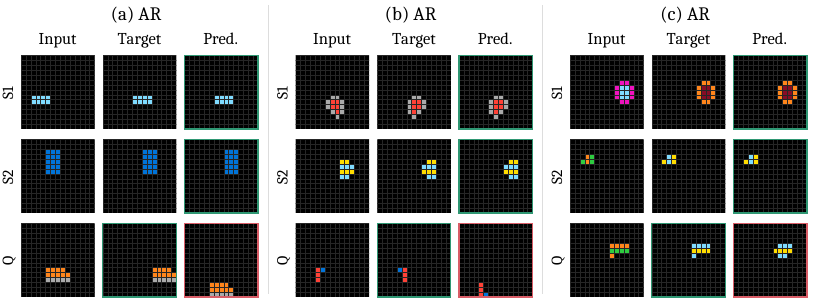}
\caption{\textbf{AR: support fit without query transfer.} Each column shows a different depth-one Single-B task from this 256-task subset (seed 42). The adapted model fits both supports but fails on the query.}
\label{fig:ar_support_transfer}
\end{figure}

We analyze saved Level-1 outputs from three seeds per method. Each set contains 10,000 tasks for shared Single, Multi-A, and Multi-B; shared Single-A/B tasks are counted once. The qualitative AR examples in Figure~\ref{fig:ar_support_transfer} are drawn from a fixed random subset of 256 tasks per mode and seed. OC and LLM outputs are excluded from these diagnostics.

All rates use full-grid exactness and report the mean $\pm$ population SD over three seeds (ddof=0). Conditional rates are computed per seed before aggregation. Every-support exactness ($S$) is a diagnostic, not an official benchmark metric. Cont-Mono, Disc-Mono, and NEO select among per-support candidates by exact support coverage, breaking ties by shorter length and then earlier support. Opt starts from the average of the latents inferred from each support pair and optimizes it to fit all supports; AR adapts on all supports. Thus conditional cohorts differ by model. These are checkpoint diagnostics, not curriculum \textsc{Score} reproductions or controlled architectural ablations.

Full-set Opt rates use archived final outputs. Depth-one and paired optimization rates use separately recorded step-10 trajectories; paired changes compare each trajectory to its own step 0, not to the cross-pair-selected Cont-Mono output.
\begin{table}[!ht]
\centering
\footnotesize
\setlength{\tabcolsep}{4pt}
\caption{\textbf{Support fit and query transfer at Lv1.} Mean $\pm$ SD across three seeds (\%). $S$: every support correct; $Q$: query correct; $Q\mid S$: query accuracy conditional on $S$. Each dataset contains 10,000 tasks per seed; Single-A/B share the same Lv1 data. See Appendix~\ref{app:failure_scope} for protocols.}
\label{tab:support_transfer_mean}
\begin{tabular}{llrrr}
\toprule
Dataset & Method & $S$ (\%) & $Q$ (\%) & $Q\mid S$ (\%) \\
\midrule
Single-A/B & Cont-Mono & 21.93 $\pm$ 1.81 & 16.66 $\pm$ 1.24 & 25.98 $\pm$ 2.32 \\
 & Cont-Mono-Opt & 23.84 $\pm$ 1.54 & 16.97 $\pm$ 1.23 & 25.51 $\pm$ 1.90 \\
 & Disc-Mono & 50.19 $\pm$ 5.33 & 54.48 $\pm$ 5.66 & 84.86 $\pm$ 5.50 \\
 & NEO & 87.01 $\pm$ 4.08 & 85.81 $\pm$ 4.17 & 92.11 $\pm$ 2.77 \\
 & AR & 99.95 $\pm$ 0.03 & 7.32 $\pm$ 0.45 & 7.31 $\pm$ 0.44 \\
\midrule
Multi-A & Cont-Mono & 10.64 $\pm$ 3.98 & 10.45 $\pm$ 4.61 & 22.74 $\pm$ 9.31 \\
 & Cont-Mono-Opt & 11.24 $\pm$ 3.82 & 10.06 $\pm$ 4.37 & 22.07 $\pm$ 9.27 \\
 & Disc-Mono & 10.87 $\pm$ 2.55 & 16.10 $\pm$ 2.37 & 51.84 $\pm$ 0.97 \\
 & NEO & 39.07 $\pm$ 2.00 & 36.66 $\pm$ 3.54 & 52.68 $\pm$ 3.39 \\
 & AR & 99.99 $\pm$ 0.01 & 7.78 $\pm$ 0.45 & 7.78 $\pm$ 0.45 \\
\midrule
Multi-B & Cont-Mono & 4.35 $\pm$ 0.47 & 5.57 $\pm$ 0.61 & 27.43 $\pm$ 1.22 \\
 & Cont-Mono-Opt & 4.95 $\pm$ 0.33 & 5.48 $\pm$ 0.60 & 25.67 $\pm$ 2.60 \\
 & Disc-Mono & 4.98 $\pm$ 0.06 & 8.47 $\pm$ 0.46 & 46.40 $\pm$ 2.48 \\
 & NEO & 21.16 $\pm$ 1.03 & 22.55 $\pm$ 1.34 & 50.89 $\pm$ 0.62 \\
 & AR & 99.92 $\pm$ 0.04 & 9.29 $\pm$ 0.87 & 9.30 $\pm$ 0.87 \\
\bottomrule
\end{tabular}
\end{table}

\begin{table}[!ht]
\centering\footnotesize
\setlength{\tabcolsep}{3pt}
\caption{\textbf{Single-A/B transfer at Lv1, primitive depth one.} Mean $\pm$ SD across three seeds (\%). $S$: every support correct; $Q$: query correct. $S\cap\neg Q$ uses all $N$ tasks as its denominator. Inference follows Appendix~\ref{app:failure_scope}.}
\label{tab:support_transfer_seed}
\begin{tabular}{lrrrr}
\toprule
Method & $N$ & $S$ (\%) & $Q$ (\%) & $S\cap\neg Q$ (\%) \\
\midrule
Cont-Mono & 2944 & 34.69 $\pm$ 2.73 & 26.18 $\pm$ 2.01 & 23.51 $\pm$ 1.04 \\
Cont-Mono-Opt & 2944 & 36.16 $\pm$ 2.20 & 26.51 $\pm$ 1.97 & 24.63 $\pm$ 0.79 \\
Disc-Mono & 2944 & 73.97 $\pm$ 7.36 & 75.51 $\pm$ 8.09 & 8.48 $\pm$ 3.36 \\
NEO & 2944 & 97.54 $\pm$ 2.29 & 96.49 $\pm$ 2.69 & 2.36 $\pm$ 1.41 \\
AR & 2944 & 99.95 $\pm$ 0.02 & 13.07 $\pm$ 0.78 & 86.92 $\pm$ 0.77 \\
\bottomrule
\end{tabular}
\end{table}

\begin{table}[!ht]
\centering\footnotesize
\setlength{\tabcolsep}{3pt}
\caption{\textbf{Single-A/B transfer at Lv1, primitive depth two.} Mean $\pm$ SD across three seeds (\%). $N$: tasks per seed; $S$: every support correct; $Q$: query correct. $S\cap\neg Q$ uses all $N$ tasks as its denominator. Inference follows Table~\ref{tab:support_transfer_seed}, including step-10 trajectories for Cont-Mono-Opt. Depth is the number of primitives in the generating program.}
\label{tab:support_transfer_depth2}
\begin{tabular}{lrrrr}
\toprule
Method & $N$ & $S$ (\%) & $Q$ (\%) & $S\cap\neg Q$ (\%) \\
\midrule
Cont-Mono & 5017 & 20.01 $\pm$ 1.83 & 15.19 $\pm$ 1.15 & 15.51 $\pm$ 0.93 \\
Cont-Mono-Opt & 5017 & 22.27 $\pm$ 1.30 & 15.59 $\pm$ 1.11 & 17.34 $\pm$ 0.56 \\
Disc-Mono & 5017 & 56.19 $\pm$ 6.70 & 61.64 $\pm$ 6.80 & 9.27 $\pm$ 2.74 \\
NEO & 5017 & 89.44 $\pm$ 4.75 & 88.22 $\pm$ 4.78 & 6.54 $\pm$ 2.24 \\
AR & 5017 & 99.95 $\pm$ 0.03 & 5.97 $\pm$ 0.33 & 93.98 $\pm$ 0.36 \\
\bottomrule
\end{tabular}
\end{table}

\noindent\textbf{Output-pattern counts.}
\label{app:failure_patterns}
A one-constituent error exactly matches either constituent applied alone on a depth-two task; this does not identify an internal omitted step. Target errors require a one-step unary task with a unique target and an exact alternative-target match, excluding no-op candidates. Their frequency also depends on primitive execution accuracy.
\begin{table}[!ht]
\centering\footnotesize
\setlength{\tabcolsep}{3pt}
\caption{\textbf{Output-pattern frequencies.} Mean $\pm$ SD across three seeds (\%). Eligible $n$: Single 5,017; Multi-A 2,177; Multi-B 2,117; full $n$: 10,000. Patterns are non-exclusive. Opt uses 2,944 depth-one tasks; full-set rescue/harm: $1.12\pm0.06$/$0.99\pm0.05$\%. Within-trajectory net gain: $0.43\pm0.11$ pp.}
\label{tab:failure_pattern_counts}
\begin{tabular}{llrrr}
\toprule
Method & Denominator & Single & Multi-A & Multi-B \\
 & & One constituent & Wrong target & Wrong target \\
\midrule
Cont-Mono & Eligible & 15.05 $\pm$ 1.15 & 8.82 $\pm$ 4.89 & 4.96 $\pm$ 0.73 \\
 & Full set & 7.55 $\pm$ 0.58 & 1.92 $\pm$ 1.07 & 1.05 $\pm$ 0.16 \\
Cont-Mono-Opt & Eligible & 15.16 $\pm$ 0.94 & 8.91 $\pm$ 5.22 & 5.23 $\pm$ 0.85 \\
 & Full set & 7.60 $\pm$ 0.47 & 1.94 $\pm$ 1.14 & 1.11 $\pm$ 0.18 \\
Disc-Mono & Eligible & 3.70 $\pm$ 0.94 & 14.91 $\pm$ 4.29 & 5.78 $\pm$ 0.22 \\
 & Full set & 1.86 $\pm$ 0.47 & 3.25 $\pm$ 0.93 & 1.22 $\pm$ 0.05 \\
NEO & Eligible & 0.55 $\pm$ 0.52 & 25.43 $\pm$ 2.04 & 15.19 $\pm$ 0.74 \\
 & Full set & 0.28 $\pm$ 0.26 & 5.54 $\pm$ 0.44 & 3.22 $\pm$ 0.16 \\
\midrule
Opt: rescue & Depth one & 3.80 $\pm$ 0.21 & -- & -- \\
Opt: harm & Depth one & 3.37 $\pm$ 0.18 & -- & -- \\
\bottomrule
\end{tabular}
\end{table}

\clearpage
\subsection{Primitive Transfer and Latent Refinement}
The following examples examine whether fitting observed pairs yields transformations that remain correct on query inputs, and how latent refinement changes that behavior.

\begin{figure}[!ht]
\centering
\includegraphics[width=\linewidth]{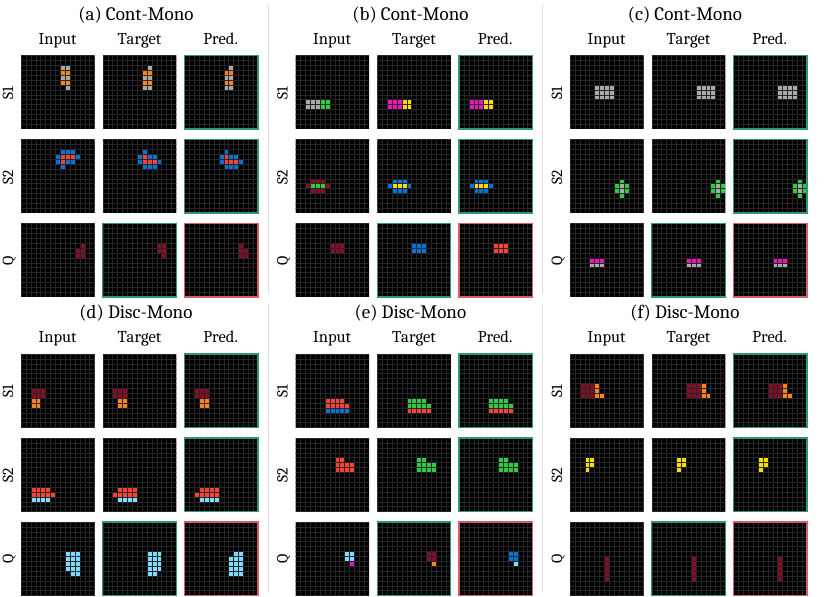}
\caption{\textbf{Transfer errors on depth-1 programs.} Cont-Mono (a--c) and Disc-Mono (d--f) fit both supports but make reflection-axis, color-change, and translation-distance errors, respectively (seed 42). Color predictions match two shifts instead of one. The Disc-Mono translation error is a rare illustrative case.}
\label{fig:single_action_transfer}
\end{figure}

\clearpage
\begin{figure}[!ht]
\centering
\includegraphics[width=\linewidth]{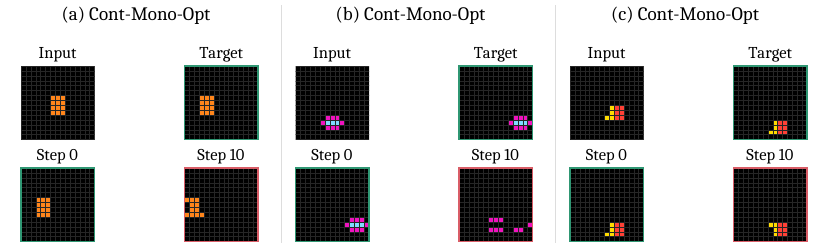}
\caption{\textbf{Support fitting can damage a correct query.} Three Cont-Mono-Opt cases change from correct at step 0 to incorrect at step 10. Both supports remain exact at recorded steps 0, 1, 5, and 10; support loss need not be constant.}
\label{fig:opt10_regressions}
\end{figure}
\label{app:opt_regressions}

\subsection{Target Selection}
The predictions below match the intended primitive applied to a different object. These output matches identify target discrepancies but do not distinguish selection-rule inference from execution errors.

\begin{figure}[!ht]
\centering
\includegraphics[width=\linewidth]{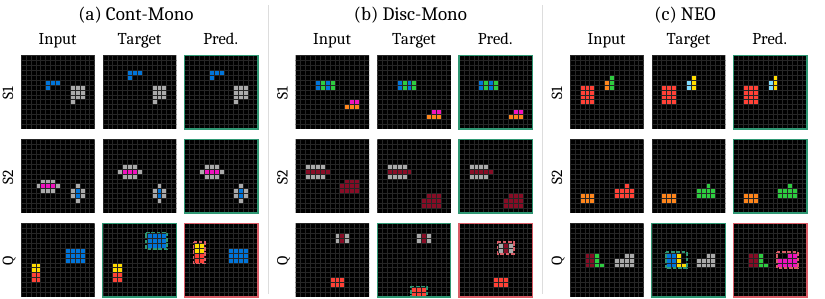}
\caption{\textbf{Correct transformation, wrong target.} Cont-Mono, Disc-Mono, and NEO fit both supports but change the wrong query object in Multi-A (seed 42). Each prediction exactly matches the correct primitive applied to another target. Dashed outlines identify the affected objects.}
\label{fig:target_transfer}
\end{figure}

\clearpage
\subsection{State-dependent execution}
\label{app:state_dependent_execution}
To test whether a learned code behaves consistently on intermediate execution states, we hold NEO weights and inferred codes fixed. Rotation and fresh reconstruction are exact on supports and query. ``Continue'' uses the existing hidden state; ``Re-encode'' re-encodes the decoded image. These selected cases do not estimate prevalence.

\begin{figure}[!ht]
\centering
\includegraphics[width=\linewidth]{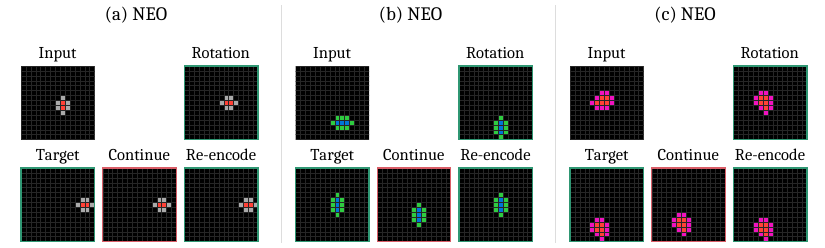}
\caption{\textbf{Re-encoding corrects some NEO continuations.} The first rotation is exact, but the following translation moves 3, 3, and 4 cells instead of 4, 5, and 5. Re-encoding the decoded image before applying the same second code corrects these cases.}
\label{fig:state_reencoding}
\end{figure}

\begin{figure}[!ht]
\centering
\includegraphics[width=\linewidth]{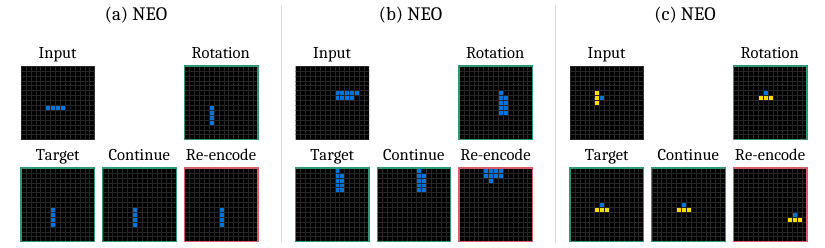}
\caption{\textbf{Re-encoding also breaks correct NEO continuations.} Under the same intervention, three previously correct queries become incorrect. Together with Figure~\ref{fig:state_reencoding}, these cases show state-dependent execution; re-encoding is not a general remedy.}
\label{fig:state_reencoding_harm}
\end{figure}

\clearpage
\section{Latent-space visualization}
\label{app:latent-visualization}

We select tasks generated by depth-1 programs from the Single-A
Level 1 dataset and provide the first support input--output pair
of each task to each model's programmer.
We visualize the resulting latent representations using t-SNE.
Each point is colored according to the primitive that generated its source task;
colors do not denote the learned semantics of the latent vectors or codebook entries.

For Cont-Mono, we project the continuous latent vectors.
For Disc-Mono and NEO, we jointly project the pre-quantization
continuous vectors and the codebook vectors.
We fit t-SNE independently for each checkpoint.
All codebook entries are included in the projection, but only entries
selected by the plotted vectors in the original latent space are displayed.

\begin{figure}[!ht]
\centering
\includegraphics[width=\linewidth]{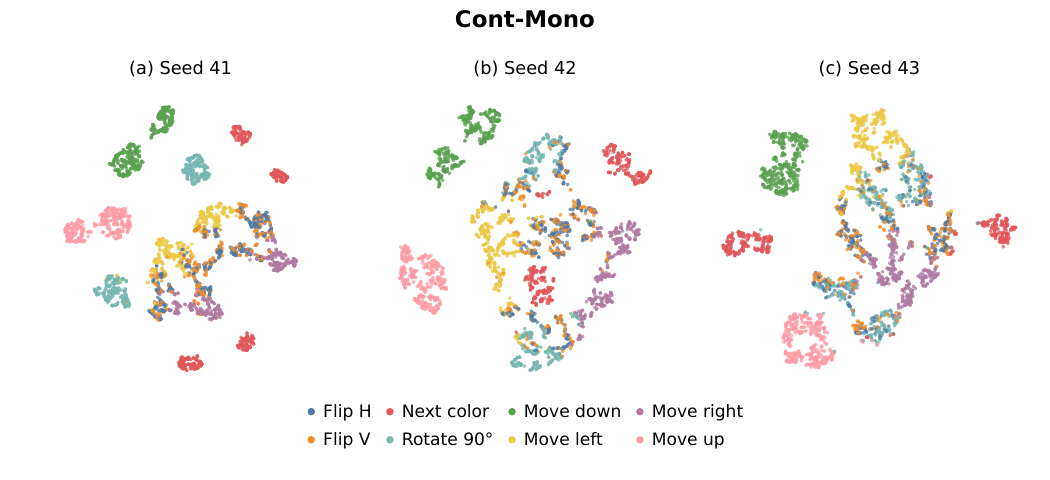}
\caption{\textbf{Continuous latent representations of Cont-Mono.}
Each point represents the latent inferred from a task's first support pair.
Colors identify the primitives that generated the source tasks.}
\label{fig:latent_cont_mono}
\end{figure}

\begin{figure}[!ht]
\centering
\includegraphics[width=\linewidth]{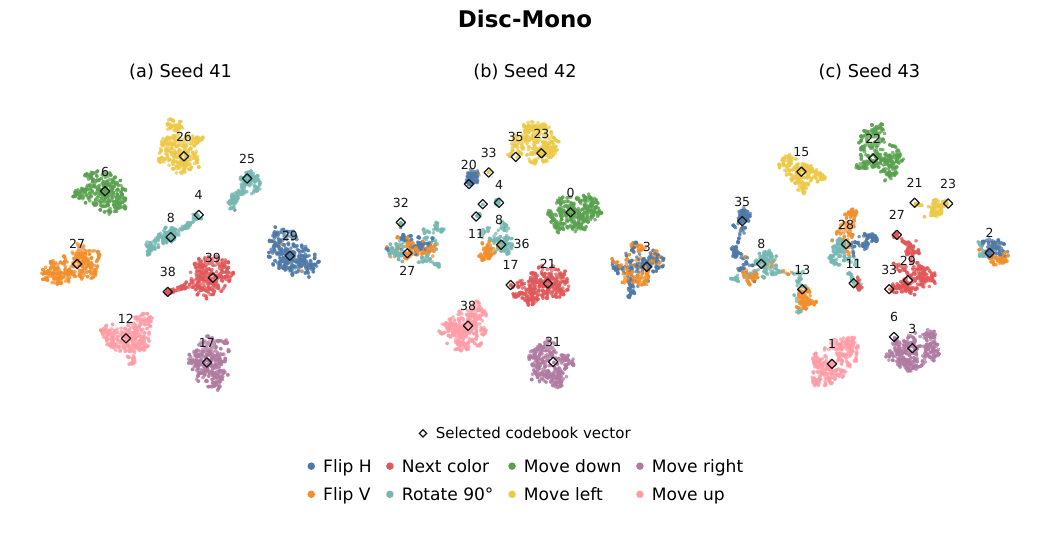}
\caption{\textbf{Pre-quantization representations and codebook vectors of Disc-Mono.}
Points represent continuous vectors before quantization,
colored by the primitives that generated the source tasks.
Hollow diamonds mark selected codebook entries, and numeric labels
denote code IDs within each checkpoint.}
\label{fig:latent_disc_mono}
\end{figure}

\begin{figure}[!ht]
\centering
\includegraphics[width=\linewidth]{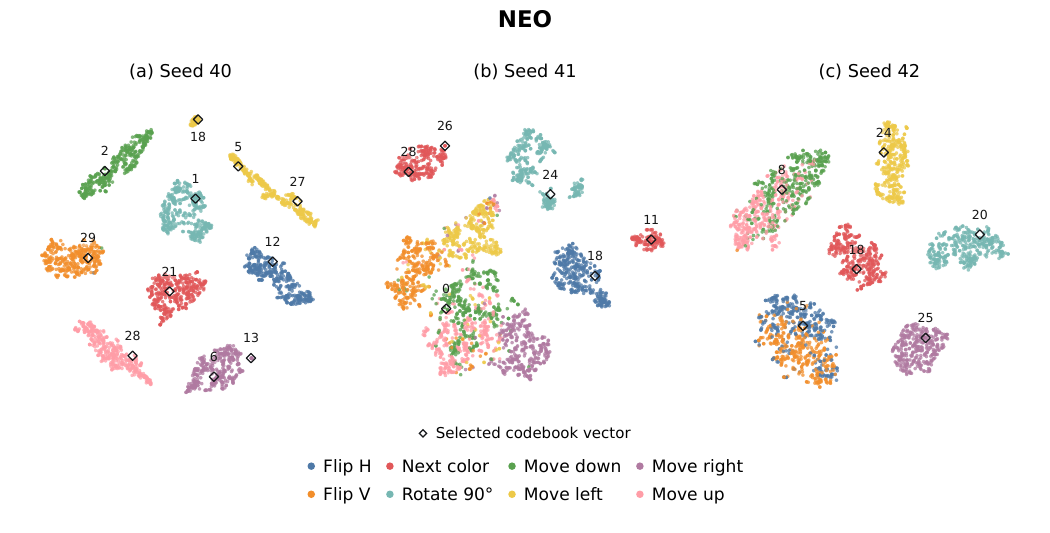}
\caption{\textbf{Pre-quantization representations and codebook vectors of NEO.}
Points represent continuous vectors before quantization,
colored by the primitives that generated the source tasks.
Hollow diamonds mark selected codebook entries, and numeric labels
denote code IDs within each checkpoint.}
\label{fig:latent_neo}
\end{figure}

\clearpage
\section{Licenses}
\label{app:licenses}

\begin{center}
\captionof{table}{\textbf{Existing assets, licenses, and source links.} We list the existing models and public code repositories used in our experiments. The BabyTheorist data and generator are created by the authors and are therefore not external assets.}
\label{tab:licenses}
\small
\begin{tabularx}{\linewidth}{>{\raggedright\arraybackslash}p{0.17\linewidth}>{\raggedright\arraybackslash}X>{\raggedright\arraybackslash}p{0.16\linewidth}>{\raggedright\arraybackslash}p{0.36\linewidth}}
\toprule
\textbf{Asset} & \textbf{Use in this paper} & \textbf{License / terms} & \textbf{Source link} \\
\midrule
GPT-OSS 120B & Open-weight LLM baseline (Section~\ref{sec:baselines}) & Apache License 2.0 & \url{https://huggingface.co/openai/gpt-oss-120b} \\
Qwen 3.5 27B & Open-weight LLM baseline (Section~\ref{sec:baselines}) & Apache License 2.0 & \url{https://huggingface.co/Qwen/Qwen3.5-27B} \\
NEO / L2T repository~\citep{baek2026l2t} & NEO baseline reference implementation (Appendix~\ref{sec:neo}) & MIT License & \url{https://github.com/ahn-ml/learning-to-theorize} \\
Dreamweaver repository~\citep{baek2025dreamweaver} & Slot Attention and object-centric components for Cont-Mono-OC & MIT License & \url{https://github.com/ahn-ml/dreamweaver-release} \\
DUO repository~\citep{sahoo2025the} & Reference implementation used in our baseline code & Apache License 2.0 & \url{https://github.com/s-sahoo/duo} \\
GPT-5.6 Sol & Proprietary reference model (Appendix~\ref{app:gpt56_sol}) & Accessed via the OpenAI API under OpenAI's terms of use & \url{https://openai.com/policies/terms-of-use} \\
\bottomrule
\end{tabularx}
\end{center}

\textbf{BabyTheorist release.} The generator and evaluation code are released under the MIT License at \href{https://github.com/ahn-ml/babytheorist}{\texttt{github.com/ahn-ml/babytheorist}}. The dataset is released under CC0~1.0 on Hugging Face at \href{https://huggingface.co/datasets/ahn-ml/babytheorist}{\texttt{huggingface.co/datasets/ahn-ml/babytheorist}} and is archived on Harvard Dataverse with a persistent DOI, \href{https://doi.org/10.7910/DVN/H9UGTL}{\texttt{doi:10.7910/DVN/H9UGTL}}, which also provides its Croissant metadata.

%% file: CameraReady/tables/tab_sec5_result.tex
\begin{table}[h]
\centering
\small
\caption{\textsc{Score} on the four BabyTheorist benchmark modes. \textbf{\textsc{Score}} sums per-level exact-match accuracy across reached levels. Higher is better. Mean over $n=3$ seeds (population std); LLM rows are single evaluations on the first 100 tasks per level.}
\label{tab:main_results}
\begin{tabularx}{\linewidth}{@{}Xcccc@{}}
\toprule
\textbf{Model} & \textbf{Single-A} & \textbf{Single-B} & \textbf{Multi-A} & \textbf{Multi-B} \\
\midrule
LLM (GPT-OSS 120B, high) & 59 & 52 & 15 & 15 \\
LLM (Qwen 3.5 27B, thinking-off) & 8 & 8 & 4 & 6 \\
AR & $6.7 \pm 0.6$ & $7.3 \pm 0.5$ & $7.7 \pm 0.6$ & $9.3 \pm 1.0$ \\
Cont-Mono & $15.9 \pm 1.4$ & $16.7 \pm 1.5$ & $10.5 \pm 5.7$ & $5.6 \pm 0.7$ \\
Disc-Mono & $52.5 \pm 7.0$ & $54.5 \pm 6.9$ & $16.1 \pm 2.9$ & $8.5 \pm 0.6$ \\
Cont-Mono-Opt & $16.8 \pm 1.4$ & $17.9 \pm 1.6$ & $8.7 \pm 4.9$ & $5.6 \pm 0.7$ \\
Cont-Mono-OC & - & - & $3.2 \pm 1.4$ & $10.7 \pm 5.6$ \\
NEO & $\mathbf{136.8 \pm 49.6}$ & $\mathbf{137.7 \pm 49.8}$ & $\mathbf{36.7 \pm 4.3}$ & $\mathbf{22.6 \pm 1.6}$ \\
\bottomrule
\end{tabularx}
\end{table}

%% file: checklist.tex
\section*{NeurIPS Paper Checklist}

\begin{enumerate}

\item {\bf Claims}
    \item[] Question: Do the main claims made in the abstract and introduction accurately reflect the paper's contributions and scope?
    \item[] Answer: \answerYes{} 
    \item[] Justification: The introduction and the abstract include the main contributions and scope: the benchmark and our experiments.

\item {\bf Limitations}
    \item[] Question: Does the paper discuss the limitations of the work performed by the authors?
    \item[] Answer: \answerYes{} 
    \item[] Justification: We discuss the limitations of our work in Section~\ref{sec:conclusion}.

\item {\bf Theory assumptions and proofs}
    \item[] Question: For each theoretical result, does the paper provide the full set of assumptions and a complete (and correct) proof?
    \item[] Answer: \answerNA{} 
    \item[] Justification: Our work does not contain theoretical results.

    \item {\bf Experimental result reproducibility}
    \item[] Question: Does the paper fully disclose all the information needed to reproduce the main experimental results of the paper to the extent that it affects the main claims and/or conclusions of the paper (regardless of whether the code and data are provided or not)?
    \item[] Answer: \answerYes{} 
    \item[] Justification: We disclose all experimental configurations in Appendix~\ref{app:hyperparam} for reproducibility. We checked that generation of the provided datasets is also reproducible with the provided dataset generation code.

\item {\bf Open access to data and code}
    \item[] Question: Does the paper provide open access to the data and code, with sufficient instructions to faithfully reproduce the main experimental results, as described in supplemental material?
    \item[] Answer: \answerYes{} 
    \item[] Justification: The dataset (Hugging Face, archived on Harvard Dataverse) and the code for data generation, the benchmark environment, and evaluation (GitHub) are linked on the first page, with a dataset card and README describing usage. Baseline architectures, hyperparameters, and training settings are described in Appendices~\ref{sec:neo} and~\ref{app:hyperparam}.

\item {\bf Experimental setting/details}
    \item[] Question: Does the paper specify all the training and test details (e.g., data splits, hyperparameters, how they were chosen, type of optimizer) necessary to understand the results?
    \item[] Answer: \answerYes{} 
    \item[] Justification: We describe the benchmark pipeline in Figure~\ref{fig:babytheorist_pipeline} and the experimental configuration in Appendix~\ref{app:hyperparam}.

\item {\bf Experiment statistical significance}
    \item[] Question: Does the paper report error bars suitably and correctly defined or other appropriate information about the statistical significance of the experiments?
    \item[] Answer: \answerYes{} 
    \item[] Justification: We report standard deviation based $1\sigma$ error bars across three runs with different seeds in Figure~\ref{fig:main_results} and Table~\ref{tab:main_results}.

\item {\bf Experiments compute resources}
    \item[] Question: For each experiment, does the paper provide sufficient information on the computer resources (type of compute workers, memory, time of execution) needed to reproduce the experiments?
    \item[] Answer: \answerYes{} 
    \item[] Justification: Compute resources and execution time for training and inference are reported in Appendix~\ref{app:hyperparam}.
    
\item {\bf Code of ethics}
    \item[] Question: Does the research conducted in the paper conform, in every respect, with the NeurIPS Code of Ethics \url{https://neurips.cc/public/EthicsGuidelines}?
    \item[] Answer: \answerYes{} 
    \item[] Justification: We confirmed this work conforms to the NeurIPS Code of Ethics in every aspect. We carefully evaluated potential harms and societal effect of our benchmark.

\item {\bf Broader impacts}
    \item[] Question: Does the paper discuss both potential positive societal impacts and negative societal impacts of the work performed?
    \item[] Answer: \answerYes{} 
    \item[] Justification: We discuss broader impacts of this work in the Broader Impact section.
    
\item {\bf Safeguards}
    \item[] Question: Does the paper describe safeguards that have been put in place for responsible release of data or models that have a high risk for misuse (e.g., pre-trained language models, image generators, or scraped datasets)?
    \item[] Answer: \answerNA{} 
    \item[] Justification: This work does not contain such risks or misuse. This work provides only purely synthetic grid-based dataset and benchmark.

\item {\bf Licenses for existing assets}
    \item[] Question: Are the creators or original owners of assets (e.g., code, data, models), used in the paper, properly credited and are the license and terms of use explicitly mentioned and properly respected?
    \item[] Answer: \answerYes{} 
    \item[] Justification: Previously developed baselines are cited in Appendix~\ref{app:baselines_long}, related benchmarks are cited in Appendix~\ref{app:related_long}, and the licenses and source links of existing assets are listed in Appendix~\ref{app:licenses}.

\item {\bf New assets}
    \item[] Question: Are new assets introduced in the paper well documented and is the documentation provided alongside the assets?
    \item[] Answer: \answerYes{} 
    \item[] Justification: The dataset is documented in its Hugging Face dataset card (structure, fields, generation, and license), and the code repository includes a README and usage examples for the benchmark environment.

\item {\bf Crowdsourcing and research with human subjects}
    \item[] Question: For crowdsourcing experiments and research with human subjects, does the paper include the full text of instructions given to participants and screenshots, if applicable, as well as details about compensation (if any)? 
    \item[] Answer: \answerNA{}  
    \item[] Justification: This paper does not incorporate crowdsourcing method nor human subjects.

\item {\bf Institutional review board (IRB) approvals or equivalent for research with human subjects}
    \item[] Question: Does the paper describe potential risks incurred by study participants, whether such risks were disclosed to the subjects, and whether Institutional Review Board (IRB) approvals (or an equivalent approval/review based on the requirements of your country or institution) were obtained?
    \item[] Answer: \answerNA{} 
    \item[] Justification: No human-subject research was performed.

\item {\bf Declaration of LLM usage}
    \item[] Question: Does the paper describe the usage of LLMs if it is an important, original, or non-standard component of the core methods in this research? Note that if the LLM is used only for writing, editing, or formatting purposes and does \emph{not} impact the core methodology, scientific rigor, or originality of the research, declaration is not required.
    \item[] Answer: \answerNA{} 
    \item[] Justification: The development of core benchmark methodology is not impacted by usage of LLMs.

\end{enumerate}